\documentclass{article}

\usepackage[main,final]{neurips_2026}

\usepackage[utf8]{inputenc} 
\usepackage[T1]{fontenc}    
\usepackage{hyperref}       
\usepackage{url}            
\usepackage{booktabs}       
\usepackage{amsfonts}       
\usepackage{nicefrac}       
\usepackage{microtype}      
\usepackage{xcolor}         
\usepackage[most]{tcolorbox}
\usepackage{enumitem}
\usepackage{float}
\usepackage{titletoc}
\usepackage{etoc}
\usepackage{amsmath}
\usepackage{placeins}
\usepackage{graphicx}
\usepackage{subcaption}
\usepackage{wrapfig}
\usepackage{algorithm}
\usepackage{algpseudocode}
\usepackage{float}
\usepackage[most]{tcolorbox}
\usepackage{enumitem}
\definecolor{TableGroup}{HTML}{F3F5F7}
\definecolor{OursRow}{HTML}{EEF4FF}
\definecolor{OursRow-RL}{HTML}{F8F3FF}
\usepackage[table]{xcolor}  

\NewDocumentCommand{\hongru}{ mO{} }{\textcolor{blue}{\textsuperscript{\textit{Hongru}}\textsf{\textbf{\small[#1]}}}}

\title{
Towards On-Policy Data Evolution for Visual-Native Multimodal Deep Search Agents
}

\author{%
  Shijue Huang$^{\alpha,\beta}$, Hangyu Guo$^{\alpha,\beta}$, Guanting Dong$^{\gamma}$, Chenxin Li$^{\delta}$ \\
  \textbf{Junting Lu$^{\epsilon}$, Xinyu Geng$^{\alpha,\beta}$, Zhaochen Su$^{\alpha,\beta}$, Zhenyu Li$^{\zeta}$} \\
  \textbf{Shuang Chen$^{\delta}$, Hongru Wang$^{\eta}$, Yi R. (May) Fung$^{\alpha,\beta}$\thanks{Corresponding author.}} \\
  $^{\alpha}$The Hong Kong University of Science and Technology, $^{\beta}$MMSense Lab \\
  $^{\gamma}$Renmin University of China, $^{\delta}$The Chinese University of Hong Kong \\
  $^{\epsilon}$Peking University, $^{\zeta}$Tsinghua University, $^{\eta}$University of Edinburgh \\
  \texttt{shijue.huang@connect.ust.hk} \quad \texttt{yrfung@ust.hk}
}

\begin{document}

\setcounter{footnote}{1}
\maketitle
\setcounter{footnote}{0}

\begin{abstract}

Multimodal deep search requires an agent to solve open-world problems by chaining search, tool use, and visual reasoning over evolving textual and visual context. Two bottlenecks limit current systems. First, existing tool-use harnesses treat images returned by search, browsing, or transformation as transient outputs, so intermediate visual evidence cannot be re-consumed by later tools. Second, training data is usually built by fixed curation recipes that cannot track the target agent's evolving capability. To address these challenges, we first introduce a visual-native agent harness centered on an \emph{image bank reference protocol}, which registers every tool-returned image as an addressable \texttt{<image:N>} reference
and makes intermediate visual evidence reusable by later tools.
On top of this harness, On-policy Data Evolution (ODE) runs a closed-loop data generator that refines itself across rounds from rollouts of the policy being trained. This per-round refinement makes each round's data target what the current policy still needs to learn. The same framework supports both diverse supervised fine-tuning data and policy-aware reinforcement learning data curation, covering the full training lifecycle of the target agent. 
Across 8 multimodal deep search benchmarks, ODE improves
the Qwen3-VL-8B agent from $24.9\%$ to $39.0\%$ on average. At 30B, ODE raises the average score from $30.6\%$ to $41.5\%$. Further analyses validate the effectiveness of image bank reuse, especially on complex tasks requiring iterative visual refinement, while rollout-feedback evolution yields more grounded SFT traces and better policy-matched RL tasks than static synthesis.\footnote{Our implementation is available at \url{https://github.com/JoeYing1019/ODE}.}

\end{abstract}

\section{Introduction}
Recently, Multimodal Large Language Models (MLLMs) have witnessed a rapid emergence of agent capabilities, pushing their application boundary from static image-question answering toward open-world deep search~\citep{openai2023gpt4v,bytedance2026seed20,qwen3vl2025,jiang2025mmsearch,li2025mmbrowsecompcomprehensivebenchmarkmultimodal,tao2026mmsearchplus}. In this emerging setting, a model is expected to interact with search engines and a broad ecosystem of external tools in real time, gathering evidence to generate grounded answers. In practice, user information needs are becoming increasingly complex and open-ended, where shallow retrieval no longer suffices to capture their intent~\citep{jiang2025mmsearch,li2025mmbrowsecompcomprehensivebenchmarkmultimodal,tao2026mmsearchplus,su2026agentvistaevaluatingmultimodalagents}. To succeed, an agent must propagate fine-grained visual cues across turns, cross-verify evidence across modalities, and sustain coherent multi-step reasoning in noisy, evolving environments. This makes multimodal deep search far more demanding, requiring not only perception but also tool use and cross-modal evidence integration over an expanding context.

Building strong multimodal deep search agents remains challenging for two reasons: \textbf{(1) Existing pipelines underutilize persistent visual state in tool-augmented search:} Early multimodal search agents augment MLLMs with image and text search to enable on-demand retrieval in open-world environments~\citep{wu2025mmsearchr1incentivizinglmmssearch,geng2026webwatcher}, and subsequent works extend this paradigm with crop-conditioned image search, iterative query refinement, and increasingly complex multi-turn visual-textual exploration~\citep{narayan2025deepmmsearchr1empoweringmultimodalllms,hong2026deepeyesv,huang2026visiondeepresearchincentivizingdeepresearchcapability}. 
However, many existing approaches still center visual reasoning and search around the original task image, rather than treating tool-produced visual outputs as new reusable evidence throughout the trajectory.
\textbf{(2) Multimodal deep search data synthesis lacks closed-loop modeling of agent search behavior.} 
Recent works mainly rely on synthetic or semi-automatically constructed data.
For instance, MMSearch-R1~\citep{wu2025mmsearchr1incentivizinglmmssearch}, DeepMMSearch-R1~\citep{narayan2025deepmmsearchr1empoweringmultimodalllms}, and WebWatcher~\citep{geng2026webwatcher} obtain training data via semi-automated VQA curation, web-grounded task synthesis, and synthetic multimodal tool-use trajectories. These efforts mark important progress, but the generation recipe is usually fixed before scaling, making it difficult to use target-agent feedback to steer data toward the policy's learning frontier.

These observations suggest that progress in multimodal deep search depends on \textit{jointly advancing the agent's interaction workspace and the way its training data are constructed.} 
Motivated by this, we seek to elicit multimodal deep search capability through a co-design along these two complementary axes. On the workspace side, instead of treating multimodal search as a fixed interaction over the original task image, we build a \textbf{Visual-Native Agent Harness} that unifies 9 core tools in a shared workspace: web search, image search, scholar search, visit (browsing), visual search (Google Lens), zoom-in, rotation, flip, and Python execution. 
At its core is an
\emph{image bank reference protocol}.
It stores the original task image and every
tool-returned image as reusable visual state, allowing later actions to operate on visual evidence
produced by earlier steps. This turns multimodal search from single-image interaction into a chained
visual workflow with evidence accumulation.

On the data side, built on our visual-native harness, we introduce \textbf{On-policy Data Evolution (ODE)}, which treats multimodal data construction as adaptive optimization rather than a fixed curation recipe. Instead of designing a synthesis pipeline once and then scaling it, ODE repeatedly generates candidate tasks, executes the target policy on them, and uses rubric-based trace analysis as feedback to revise the next round of data synthesis. In this sense, the rubric plays a role analogous to a \textit{loss function}: it identifies whether the generated data is too easy, too brittle, insufficiently visual, poorly grounded, or otherwise misaligned with the agent's current training needs. 
The same evolution principle supports both supervised fine-tuning (SFT) and reinforcement learning (RL) with mode-specific objectives: ODE favors grounded, tool-effective, and diverse teacher trajectories for SFT, and seeks verifiable tasks near the policy's learning frontier for RL.

Experiments across eight challenging multimodal deep search benchmarks spanning MMBC, HLE-VL, BC-VL, MMSearch, VDR, MMSearch+, SimpleVQA, and FVQA show that the proposed framework substantially improves agent performance at both 8B and 30B scales when evaluated under the same visual-native harness. Further controlled analyses show that both parts of the framework matter: removing reusable image bank references weakens performance most on tasks that activate secondary image use, while replacing ODE with a static synthesis recipe yields lower SFT and RL gains under matched data budgets.

 To summarize, our contributions are as follows:
  \begin{itemize}[leftmargin=1.4em,itemsep=2pt,topsep=2pt]
      \item We introduce a \textbf{Visual-Native Agent Harness} for multimodal deep search, where search, browsing and
  visual manipulation operate over an image bank reference protocol that makes tool-produced visual evidence persistently reusable across the trajectory.
      \item We propose \textbf{On-policy Data Evolution (ODE)}, a closed-loop data construction framework that couples
  task synthesis, policy rollout, rubric-based trace analysis, and configuration optimization, and supports
  both SFT-style teacher-trace curation and RL training data generation.
      \item We validate the framework across eight multimodal deep search benchmarks. ODE improves Qwen3-VL from $24.9\%$ to $39.0\%$ at 8B and from $30.6\%$ to $41.5\%$ at 30B on average, verifying the effectiveness of visual-state reuse and data evolution against static synthesis.

  \end{itemize}

\section{Method}
\label{sec:method}
\begin{figure}[t]
    \centering
    \includegraphics[width=\linewidth]{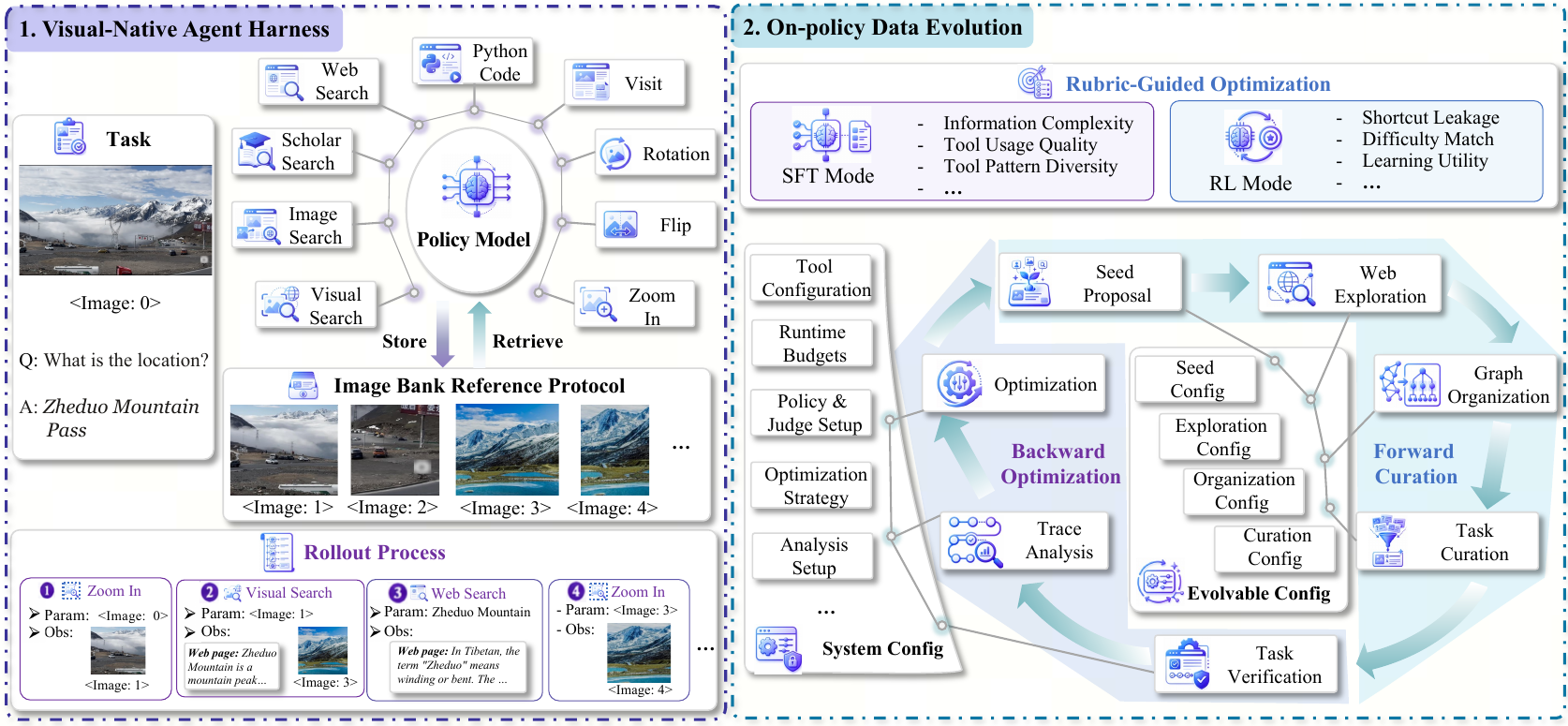}
    \caption{
    Overview of our framework.
  \textbf{Left:} The visual-native agent harness unifies 9 tools in a shared workspace and enables reusable
  visual state through the image bank reference protocol.
  \textbf{Right:} ODE constructs data with a closed loop over the harness: the forward pipeline synthesizes grounded tasks, and the backward pipeline uses rollout traces to refine the next generation configuration.
    }
    \label{fig:main}
\end{figure}

\textbf{Overview.} 
In this section, we present the proposed framework, as illustrated in Fig.~\ref{fig:main}. To further unlock the multimodal deep search agent's ability, we first propose the visual-native agent harness (Section~\ref{sec:harness}), which lets the agent reuse
intermediate visual evidence through an image bank reference protocol that registers tool-returned
images and makes them available to subsequent tool calls. 
Then, unlike static data-synthesis approaches, we propose On-policy Data Evolution (ODE, Section~\ref{sec:ode}), a closed-loop data construction process that treats the
data generator as the object being optimized.
In each ODE epoch, ODE first uses the current generator configuration to synthesize a set of candidate tasks. The target policy then rolls out these tasks in the visual-native harness, producing interaction traces of tool calls, observations, and final answers. A mode-specific rubric scores these traces and revises the generator configuration, so that the next ODE epoch produces data better aligned with the SFT or RL objective. The data generator therefore evolves with policy feedback round by round, rather than being fixed by a static curation recipe.

\subsection{Visual-Native Agent Harness}
\label{sec:harness}

Multimodal deep search often requires iterative search, browsing, visual manipulation, and computation
before answering. However, existing approaches typically organize visual reasoning around the original
task image, so tool-returned images are not consistently exposed as reusable inputs to later tools. As a
result, visual evidence cannot propagate across tool calls as naturally as textual evidence. To address this, our visual-native harness introduces an \emph{image bank reference protocol}, shown in
Fig.~\ref{fig:main} (left), which assigns each initial or tool-returned image an addressable
\texttt{\string<image:N\string>} reference and makes it available to subsequent tool calls.

Specifically, we represent a multimodal deep search task handled by the harness as $\mathcal{T} = (q, \mathcal{I}, a)$, where $q$ is an open-world multimodal query that requires the agent to gather evidence and reason across modalities, $\mathcal{I}$ is the initial visual context loaded into the image bank, and $a$ is the reference answer for verification. Each image in the bank carries an identifier of the form \texttt{\string<image:N\string>} that later tool calls reference. Starting from $(q, \mathcal{I})$, the policy model uses the nine tools shown in Fig.~\ref{fig:main}, covering web and scholarly retrieval, image and visual search, source browsing, image transformation, and Python-based computation. The Rollout Process in Fig.~\ref{fig:main} illustrates this for the question \emph{``What is the location?''}: the agent calls \texttt{zoom\_in} on the input photo \texttt{\string<image:0\string>} to crop a mountain region into \texttt{\string<image:1\string>}, runs \texttt{visual\_search} on \texttt{\string<image:1\string>} to retrieve a candidate name with a clearer photo \texttt{\string<image:3\string>}, follows up with \texttt{web\_search} to verify the candidate, and zooms into \texttt{\string<image:3\string>} to decide the final answer is \emph{``Zheduo Mountain Pass''}.

\subsection{On-policy Data Evolution}\label{sec:ode}

\subsubsection{Forward Curation}
\label{sec:forward}

Building on the visual-native harness above, ODE represents the data generator with two configuration objects: a fixed \emph{System Config}, which defines the execution environment and evaluation protocol, and an editable \emph{Evolvable Config} $\mathcal{C}_t$, which carries the generator parameters adapted from rollout feedback across rounds.
ODE initializes $\mathcal{C}_0$ by setting the seed, exploration, organization, curation, and optimization-strategy components. After round $t$, backward refinement revises these components to produce $\mathcal{C}_{t+1}$. 
We next illustrate the four forward stages controlled by $\mathcal{C}_t$, which together turn open-world
evidence into a verifiable multimodal deep search task.

\textbf{Seed Proposal.}
The seed proposer comes up with seeds, each consisting of an entity together with an associated image, that anchor downstream exploration. 
We also establish a balanced sampling mechanism to guide seed proposal across 11 topical domains, 4
capability-requirement profiles spanning perception-only, perception+search, perception+reasoning, and
perception+search+reasoning tasks, and 4 difficulty levels.
After dropping duplicates from earlier rounds, an LLM judge accepts a seed only if its image carries visual evidence such as labels, numbers, or dates, and its entity can be cross-checked against at least two independent web sources. This binds each image to a stable real-world entity and ensures that downstream tasks are grounded in verifiable evidence.

\textbf{Web Exploration.}
For each retained seed, the explorer uses the harness's 9 tools to gather supporting evidence and organizes it into \emph{anchors}, each focused on one related sub-topic such as an official source page or a map image returned by visual search. Each anchor records: (i) a textual, visual, or numerical fact,
(ii) its source URL, (iii) any tool-returned image handle in the Image Bank, and (iv) its relation to the seed or to other anchors. 
The Exploration Config in $\mathcal{C}_t$ specifies the total and image-bearing anchor budgets.

\textbf{Graph Organization.}
The graph organizer assembles the collected nodes for each seed into a multimodal evidence graph
$G$. Each node contains text, image, or tool-derived observations, and edges capture source links,
entity or event relations, and cross-modal dependencies.
The organizer further enriches $G$ with two kinds of new nodes. \emph{Reasoning nodes} are produced by running \texttt{python\_code} and \texttt{visit} over related observations to compute quantities or cross-verify facts across sources, and \emph{perception nodes} are produced by running \texttt{zoom\_in}, \texttt{rotation}, \texttt{flip}, and \texttt{visual\_search} on existing images to expose fine-grained visual details. These enrichments make derived relations, calculations, and visual details explicit in $G$, providing richer evidence for task curation.

\textbf{Task Curation.}
The curator selects a connected evidence cluster from $G$ and turns it into a verifiable candidate task
$(q,\mathcal{I}_0,a) \in \mathcal{D}_t^{\mathrm{cand}}$, together with auxiliary annotations such as planned reasoning steps, capability requirements, and difficulty. It then refines the candidate question to increase reasoning depth, for example by requiring additional evidence, hiding obvious entity names,
or removing shortcut clues while preserving the ground-truth answer. For difficulty control, the weights
in $\mathcal{C}_t$ determine the curator's preference over difficulty levels, allowing backward
refinement to adjust the difficulty distribution in later ODE epochs. Before a task enters
$\mathcal{D}_t^{\mathrm{cand}}$, validation checks that its image references resolve, its answer is
unambiguous, and its question does not reveal the intended tool-use path.

\subsubsection{Backward Optimization}
\label{sec:backward}
Backward optimization evaluates whether the candidate tasks produced by forward exploration are useful
for training and how the generator should change in the next ODE epoch. As shown in
Fig.~\ref{fig:main}, it first verifies each task by rolling out the target policy in the visual-native
harness, then analyzes the resulting traces, and finally uses rubric-guided optimization to update the
generation configuration.

\textbf{Task Verification.}  Each candidate $x_i=(q_i,\mathcal{I}_{0,i},a_i) \in \mathcal{D}_t^{\mathrm{cand}}$ is executed in the visual-native harness by the rollout model $m_t$. For SFT, $m_t$ is a teacher model, and successful rollouts provide candidate demonstrations for distillation. For RL, $m_t$ is the current policy, so the
rollout measures whether the task is appropriate for the policy that will train on it. The execution
produces a trace $\tau_i$ containing the message history, Image Bank references, final answer, and
judge outcome.

\textbf{Trace Analysis.}
Trace Analysis evaluates each rollout trace $\tau_i$ together with the construction record of
$x_i$, including the seed image, explored sources, evidence graph, and task annotations. It returns a
diagnosis $\delta_i$ containing rubric scores and stage-level fault localization. The shared rubric
dimensions assess whether the task is information-rich, visually grounded, non-leaky, and objectively
verifiable. In addition, each mode has its own training-utility dimensions. The SFT rubric adds
Step Appropriateness, Tool Usage Quality, and Tool Pattern Diversity to evaluate whether a trace is
suitable as a teacher demonstration. The RL rubric adds Capability Requirement, Difficulty Match,
and Learning Utility to evaluate whether a task provides a useful policy-optimization signal.

The diagnosis also routes each failure mode to the stage responsible for repair. Weak seed images or
entity--image mismatch are attributed to Seed Proposal; topic drift or unsupported sources to Web
Exploration; missing computations or visual transformations to Graph Organization; and leaked,
ambiguous, or miscalibrated questions to Task Curation.

\textbf{Rubric-Guided Optimization.}
The final optimization stage aggregates the per-trace diagnoses $\delta_i$ into an epoch-level
signal $\Delta_t$ for updating the data generator. The goal is to make the next ODE epoch better
match the mode-specific rubric objectives, rather than merely increasing rollout success on the current candidates.
Concretely, $\Delta_t$ edits $\mathcal{C}_t$ into $\mathcal{C}_{t+1}$ by modifying the stage-specific
Seed, Exploration, Organization, and Curation configs, together with the Optimization Strategy. Low
seed grounding shifts the Seed Config toward entities with stronger image evidence and source
support. Sparse or repetitive evidence adjusts the Exploration Config by changing search breadth,
phase depth, or the share of image-bearing anchors. Missing computations or visual operations update
the Organization Config with additional reasoning or perception enrichment. Leakage, ambiguous
answers, or off-target difficulty revise the Curation Config through difficulty weights, enhancement
prompts, and validation constraints. The Optimization Strategy records these stage-local edits together
with aggregate rubric scores, recent score trends, batch health, tail failures, SFT diversity, and RL
policy pass rate.
The next forward pass uses $\mathcal{C}_{t+1}$, whose rollouts are analyzed again to produce
$\mathcal{C}_{t+2}$ when the policy exposes new failure modes. 
Through this continued
iteration, ODE moves SFT data toward diverse, high-quality demonstrations and RL data toward
verifiable tasks at the policy's learning frontier.

We provide a full worked example of the ODE pipeline in Appendix~\ref{app:ode_details}, including the round configuration, forward generation stages, rollout verification, trace analysis, rubric-guided optimization, and consecutive configuration updates across two ODE epochs.

\begin{wrapfigure}[26]{r}{0.45\textwidth}
    \centering
    \begin{subfigure}[t]{\linewidth}
        \centering
        \includegraphics[width=\linewidth,trim=0bp 4bp 0bp 10bp,clip]{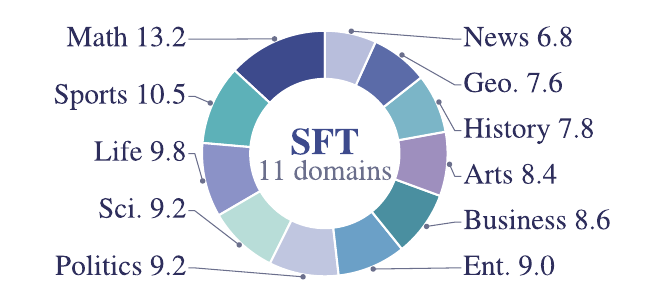}
        \caption{}
        \label{fig:domain_sft}
    \end{subfigure}

    \begin{subfigure}[t]{\linewidth}
        \centering
        \includegraphics[width=\linewidth]{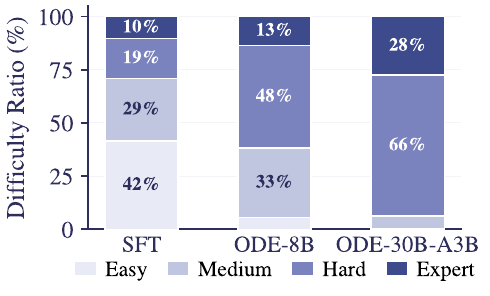}
        \caption{}
        \label{fig:difficulty}
    \end{subfigure}
    \caption{\textbf{Statistics of ODE-curated data.}
    \textbf{(a)} Topic domain coverage of the SFT demonstration set, in \%.
    \textbf{(b)} Curator-annotated difficulty ratio across the three datasets.}
    \label{fig:ode_stats}
\end{wrapfigure}

\subsection{Statistics of ODE-Curated Data}
\label{sec:data_stats}

Figure~\ref{fig:ode_stats} reports topic domain coverage and curator-annotated difficulty for the SFT training set and the two policy-specific RL task sets ODE-8B and ODE-30B-A3B curated by ODE for backbones of two different sizes, with the per-domain breakdown of the two RL sets and the planned reasoning-step distribution given in Appendix~\ref{app:data_stats_extra}.

\textbf{Topical breadth is preserved.} 
The SFT demonstration set covers all eleven topical domains, as shown in Fig.~\ref{fig:domain_sft}.
The ODE-8B and ODE-30B-A3B task sets cover the same domains, and their per-domain distributions
remain balanced, with coefficients of variation around $0.05$. Thus, policy-specific data evolution
does not collapse topical coverage while adapting task difficulty.

\textbf{Difficulty tracks policy capability.} 
The Hard plus Expert share rises from $29.06\%$ on the SFT set to $61.85\%$ on ODE-8B and
$93.67\%$ on ODE-30B-A3B, while Easy tasks fall from $41.54\%$ to $0.38\%$ over the same
progression, as shown in Fig.~\ref{fig:difficulty}. The pass-rate and difficulty-match signals collected
by the analyzer in Section~\ref{sec:backward} push the curator toward each policy's learning frontier,
so stronger policies receive proportionally harder tasks and almost no trivially easy ones.

\section{Experiments}

\subsection{Experimental Setup}
\textbf{Datasets.} We evaluate our approach on 8 multimodal deep search and related multimodal reasoning benchmarks: MM-BrowseComp (MMBC)~\citep{li2025mmbrowsecompcomprehensivebenchmarkmultimodal}, HLE-VL~\citep{phan2025lastexam}, BC-VL~\citep{geng2026webwatcher}, VDR~\citep{zeng2026visiondeepresearchbenchmarkrethinkingvisual},
MMSearch~\citep{jiang2025mmsearch},
MMSearch+~\citep{tao2026mmsearchplus}, SimpleVQA (SVQA)~\citep{cheng2025simplevqamultimodalfactualityevaluation},
  and FVQA~\citep{wang2017fvqafactbasedvisualquestion}. Details of these benchmarks are provided in
Appendix~\ref{app:benchmark_details}.

\textbf{Baselines.} We compare against proprietary and open-source multimodal models and agents under three evaluation
  settings. In the \emph{Direct Reasoning} setting, models answer in a single pass without external retrieval
  or tool use. This group includes GPT-5~\citep{singh2026openaigpt5card},
  Claude Sonnet 4~\citep{anthropic2025claude4systemcard}, Claude 3.7 Sonnet~\citep{anthropic2025claude37sonnet},
  Gemini-2.5 models~\citep{comanici2025gemini25pushingfrontier}, and the Qwen3-VL-8B-Instruct
  and Qwen3-VL-30B-A3B-Instruct backbones~\citep{qwen3vl2025}. In the \emph{Agent Workflow}
  setting, models are equipped with a general multimodal deep search toolset, including web search,
  webpage browsing, image search, and image manipulation, following prior work~\citep{huang2026visiondeepresearchincentivizingdeepresearchcapability,narayan2025deepmmsearchr1empoweringmultimodalllms}. They are prompted to solve tasks
  through iterative reasoning and tool use. We also compare with recent dedicated multimodal
  deep search agents, including MMSearch-R1~\citep{wu2025mmsearchr1incentivizinglmmssearch}
  and WebWatcher~\citep{geng2026webwatcher}. 
  For training, we instantiate our framework with two Qwen3-VL backbones:
  Qwen3-VL-8B-Instruct and Qwen3-VL-30B-A3B-Instruct~\citep{qwen3vl2025}. We refer to them as
  Qwen3-VL-8B and Qwen3-VL-30B for brevity. Further details on data construction, training, and
  evaluation are provided in Appendices~\ref{app:data}, \ref{app:train_setup}, and \ref{app:eval_setup}.

  \begin{table*}[!t]
  \centering
  \caption{Main results. Avg denotes the average score over 8 benchmarks. $\Delta$ denotes the improvement over base model.
  The best results are highlighted in \textbf{bold}, and the second-best results are \underline{underlined}.
  }
  \resizebox{\textwidth}{!}{%
  \begin{tabular}{l|ccccccccc}
  \toprule[1.5pt]
  \textbf{Model} & \textbf{MMBC} & \textbf{HLE-VL} & \textbf{BC-VL} & \textbf{VDR} & \textbf{MMSearch} & \textbf{MMSearch+} & \textbf{SVQA} & \textbf{FVQA} & \textbf{Avg.} \\
  \midrule

  \multicolumn{10}{c}{\emph{Direct Reasoning}} \\
  \midrule
  GPT-5  & 11.2 & 15.8 & 42.9 & 10.8 & 40.7 & 14.0 & 66.0 & 56.3 & 32.2 \\

  Claude-4-Sonnet & -& - & 29.3 & 2.0 & 18.7 & 4.0 & - & 35.3 & - \\
  Claude-3.7-Sonnet & -& - & 32.3 & 4.6 & 21.1 & 4.0 & - & 36.7 & - \\
  Gemini-2.5 Pro & 10.3 & \textbf{19.0} & 43.1 & 8.0 & 39.8 & 14.5 & \underline{72.7} & 60.7 & 33.5 \\
  Gemini-2.5 Flash & 4.9 & 12.0 & 37.1 & 6.2 & 30.4 & 8.1 & 67.0 & 47.7 & 26.7 \\
  Qwen3-VL-8B & 4.0 & 6.1 & 25.1 & 2.8 & 15.2 & 3.2 & 42.7 & 28.0 & 15.9 \\
  Qwen3-VL-30B & 4.5 & 5.0 & 29.6 & 3.8 & 18.7 & 3.2 & 47.0 & 34.7 & 18.3 \\

  \midrule

  \multicolumn{10}{c}{\emph{Agent Workflow}} \\
  \midrule
  GPT-5 & -

  & \underline{18.1} & \textbf{57.6} & 17.6 & 62.7 & \textbf{28.5} & 67.3 & 62.0 & -
  \\
  Claude-4-Sonnet & -& - & 48.6 & 13.6 & \underline{67.2} & 23.1 & - & \underline{69.0} & - \\
  Claude-3.7-Sonnet & -& - & \underline{50.4} & \textbf{27.2} & 63.7 & 17.2 & - & 67.3 & - \\
  Gemini-2.5 Pro & \textbf{13.8} & 17.3 & 42.3 & 10.0 & 55.7 & 24.9 & \textbf{74.3} & 65.0 & 37.9 \\
  Gemini-2.5 Flash & 11.6 & 8.5 & 38.1 & 7.8 & 53.3 & 14.5 & 67.3 & 54.7 & 32.0 \\
  Qwen3-VL-8B & 6.2 & 5.0 & 24.3 & 5.0 & 48.7 & 9.0 & 51.3 & 45.3 & 24.4 \\
  Qwen3-VL-30B & 6.2 & 4.4 & 29.3 & 4.4 & 49.3 & 6.8 & 53.0 & 45.3 & 24.8 \\
  \midrule

  \multicolumn{10}{c}{\emph{Multimodal Deep Search Agent}} \\
  \midrule
  MMSearch-R1-7B & - & - & - & - & 53.8 & - & 57.4 & 58.4 & - \\
  WebWatcher-7B & - & 10.6 & 20.3 & - & 49.1 & - & 54.3 & - & - \\
  WebWatcher-32B & - & 13.6 & 26.7 & - & 55.3 & - & 59.0 & - & - \\
  \midrule

  \multicolumn{10}{c}{\emph{Visual-Native Agent Harness (Ours)}} \\
  \midrule
  Qwen3-VL-8B & 7.6 & 6.1 & 26.1 & 4.2 & 47.3 & 10.0 & 53.0 & 44.7 & 24.9  \\
  \rowcolor{OursRow}
  + ODE-SFT & 8.5 & 8.2 & 39.6 & 19.2 & 61.7 & 21.3 & 68.3 & 62.0 & 36.1 \\
  \rowcolor{OursRow-RL}+ ODE-RL & \underline{12.5} & 11.4 & 41.9 & 20.4 & 66.0 & 24.9 & 70.3 & 64.7 & 39.0 \\
  \rowcolor{OursRow-RL}
 \hspace{2em}$\Delta$   & \textcolor{red!70!black}{+4.9} & \textcolor{red!70!black}{+5.3} & \textcolor{red!70!black}{+15.8} & \textcolor{red!70!black}{+16.2} & \textcolor{red!70!black}{+18.7} & \textcolor{red!70!black}{+14.9} & \textcolor{red!70!black}{+17.3} & \textcolor{red!70!black}{+20.0} & \textcolor{red!70!black}{+14.1} \\
  Qwen3-VL-30B & 7.1 & 8.5 & 32.8 & 11.0 & 57.0 & 10.0 & 63.7 & 54.3 & 30.6\\
  \rowcolor{OursRow}
  + ODE-SFT & 10.3 & 9.9 & 43.1 & 24.0 & 65.7 & 26.7 & 70.3 & 66.3 & \underline{39.5} \\
  \rowcolor{OursRow-RL}
  + ODE-RL & 11.2 & 10.5 & 46.1 & \underline{26.4} & \textbf{69.7} & \underline{28.1} & 71.0 & \textbf{69.3} & \textbf{41.5} \\
  \rowcolor{OursRow-RL}
\hspace{2em}$\Delta$ & \textcolor{red!70!black}{+4.1} & \textcolor{red!70!black}{+2.0} & \textcolor{red!70!black}{+13.3} & \textcolor{red!70!black}{+15.4} & \textcolor{red!70!black}{+12.7} & \textcolor{red!70!black}{+18.1} & \textcolor{red!70!black}{+7.3} & \textcolor{red!70!black}{+15.0} & \textcolor{red!70!black}{+10.9} \\
  \bottomrule[1.5pt]
  \end{tabular}
  }
  \label{tab:main}
  \end{table*}
\subsection{Main Results}
Tab.~\ref{tab:main} reports the main results. 
Three-run inference statistics for the final RL checkpoints are reported in Appendix~\ref{app:inference_variance}.
We highlight the following observations.

\textbf{(1) ODE catalyzes multimodal deep search capability.}
Under the same visual-native harness, ODE improves the Qwen3-VL-8B agent from $24.9\%$ to $39.0\%$ average accuracy, and the Qwen3-VL-30B agent from $30.6\%$ to $41.5\%$. The gains are not uniform score inflation: they are largest on benchmarks that require iterative evidence gathering and cross-modal grounding, such as VDR, MMSearch, MMSearch+, and FVQA. This suggests that ODE mainly improves the agent's ability to search, inspect, and integrate multimodal evidence over multiple steps.

\textbf{(2) Tool access is not tool competence.}
Equipping Qwen3-VL backbones with a standard agent workflow improves over direct answering, but these tool-using baselines remain far below agents trained on ODE-curated data. This indicates that multimodal deep search is not unlocked by tool access alone: the model must learn when to search, when to inspect visual evidence, how to chain tools, and how to synthesize evidence into a grounded answer. ODE addresses this at the data level by curating trajectories that demonstrate these behaviors, so that subsequent SFT and RL
optimize the model on the desired interaction patterns rather than relying on inference-time prompting alone.

\textbf{(3) Reusable state strengthens the harness.}
Before ODE training, replacing the standard agent workflow with our visual-native harness already improves the Qwen3-VL-30B-Instruct agent's average accuracy across the eight benchmarks from $24.8\%$ to $30.6\%$. The largest relative improvements appear on visually grounded and search-intensive benchmarks such as HLE-VL, VDR and MMSearch+. This supports the core harness design: tool-produced images should not be treated as transient observations, but as persistent visual state that enables
multi-step evidence construction.

\subsection{Visual-Native Harness Ablation}
\begin{figure}[t]
    \centering
    \includegraphics[width=\linewidth]{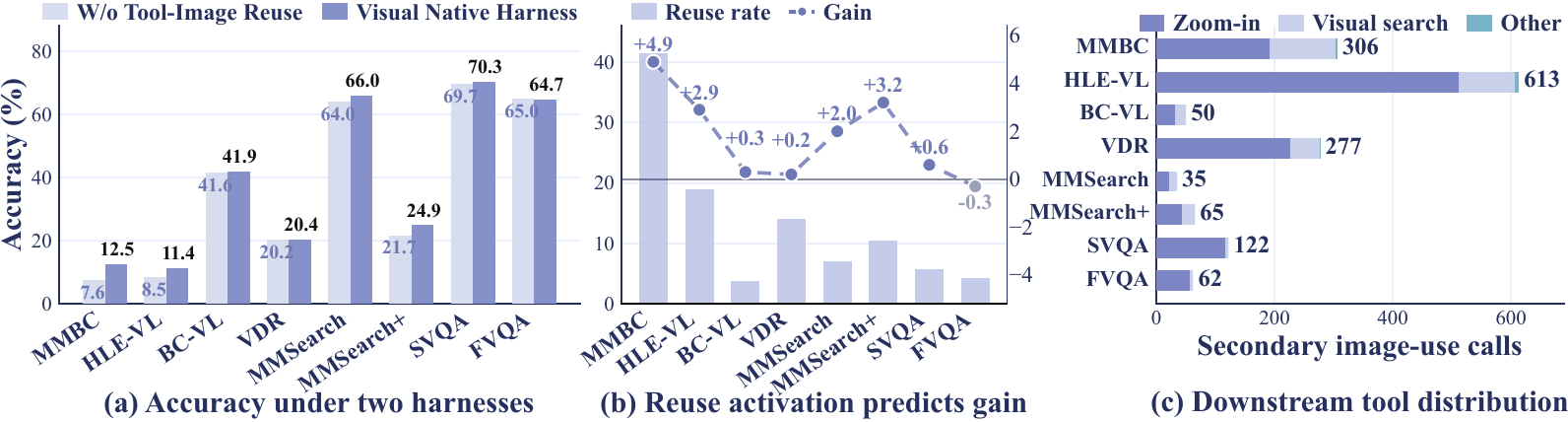}
    \caption{Visual-native harness ablation on ODE-8B-RL.}
    \label{fig:harness_analysis}
\end{figure}

This analysis evaluates the effectiveness of the proposed visual-native agent harness.
We compare ODE-8B-RL under two harnesses. The full harness keeps every tool-returned image as an addressable \texttt{<image:N>} reference, allowing later tools to consume it. The ablated harness still shows tool-returned images to the model, but removes these reusable references, so intermediate images cannot be passed into later image-consuming tools. As shown in Fig.~\ref{fig:harness_analysis}, we make the following observations:

\textbf{Reusable visual state improves performance.}
Fig.~\ref{fig:harness_analysis}(a) shows that the full visual-native harness outperforms the ablated harness on key benchmarks. The effect is especially clear on MMBC, HLE-VL, and MMSearch+, where accuracy improves by $+4.9\%$, $+2.9\%$, and $+3.2\%$, respectively. Since the ablation keeps tool-returned images visible but removes their reusable references, these gains isolate the value of making intermediate visual evidence actionable across tool calls.

\textbf{Reuse activation explains the gains.}
Fig.~\ref{fig:harness_analysis}(b) shows that benchmarks with higher secondary image-reuse rates tend to
benefit more from the full harness. MMBC has the highest reuse rate and also the largest gain, while HLE-VL and MMSearch+ show the same trend. This supports the intended interpretation of the ablation: the performance gap comes from making intermediate images reusable as later tool inputs, rather than from image visibility alone.

\textbf{Reused images support visual refinement.}
Fig.~\ref{fig:harness_analysis}(c) shows that reused images are mainly consumed by zoom-in and visual search. This indicates that the harness enables tool-produced images to be inspected, cropped, and searched again in later steps. In other words, image bank reuse turns intermediate visual outputs into working evidence for subsequent tool use.

\begin{figure}[t]
\centering
\begingroup
\fontsize{9}{10.5}\selectfont
\begin{minipage}[t]{0.70\textwidth}
  \begin{minipage}[b][1.80in][t]{\linewidth}
    \centering
    \fontsize{8.5}{9}\selectfont
    \setlength{\tabcolsep}{1.5pt}
    \renewcommand{\arraystretch}{1.13}
    \begin{tabular*}{\linewidth}{@{\extracolsep{\fill}}lrrrrr@{}}
      \toprule
      Benchmark & \shortstack{Gemini 2.5\\Flash} & \shortstack{Gemini 2.5\\Pro} & GPT-5 & \shortstack{ODE\\8B-RL} & \shortstack{ODE\\30B-A3B-RL} \\
      \midrule
      MMBC & 8.04 & 12.50 & 23.20 & 12.5 & 11.16 \\
      HLE-VL & 10.53 & 19.30 & 19.90 & 11.4 & 10.53 \\
      BC-VL & 39.10 & 45.10 & 58.60 & 41.9 & 46.10 \\
      VDR & 18.80 & 26.40 & 23.40 & 20.4 & 26.40 \\
      MMSearch & 59.33 & 61.00 & 68.00 & 66.0 & 69.67 \\
      MMSearch+ & 17.65 & 29.41 & 31.20 & 24.9 & 28.05 \\
      SVQA & 69.67 & 74.67 & 68.30 & 70.3 & 71.00 \\
      FVQA & 61.33 & 69.33 & 62.70 & 64.7 & 69.33 \\
      \midrule
      Avg. & 35.55 & 42.21 & 44.41 & 39.0 & 41.53 \\
      \bottomrule
    \end{tabular*}
  \end{minipage}
  \par
  \centering\textbf{(a) Matched-harness comparison}
\end{minipage}\hfill
\begin{minipage}[t]{0.27\textwidth}
  \begin{minipage}[b][1.80in][t]{\linewidth}
    \includegraphics[width=\linewidth]{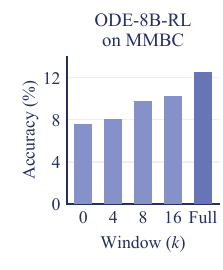}
  \end{minipage}
  \par
  \centering\textbf{(b) Image bank window}
\end{minipage}
\endgroup
\caption{Further analysis of the visual-native harness.}
\label{fig:matched_harness_window}
\end{figure}

\subsection{Further Analysis of the Visual-Native Harness}
To further assess the visual-native harness, we compare commercial models across workflows and use their performance under our harness as the basis for comparison with ODE (Fig.~\ref{fig:matched_harness_window}(a)). We then test whether restricting image reuse to recent tool outputs affects accuracy (Fig.~\ref{fig:matched_harness_window}(b)).

\textbf{Matched-harness comparison.}
Switching from the standard agent workflow (Table~\ref{tab:main}) to our visual-native harness raises average accuracy across eight benchmarks from 32.0\% to 35.55\% for Gemini 2.5 Flash and from 37.9\% to 42.21\% for Gemini 2.5 Pro. GPT-5 scores higher on all seven benchmarks reported under both workflows. These gains demonstrate the effectiveness of our harness for multimodal search across the three commercial models tested. Under the same harness, our trained ODE-30B-A3B-RL is also competitive, averaging 41.53\%, just 0.68 percentage points below Gemini 2.5 Pro.

\textbf{Image bank window.}
We evaluate ODE-8B-RL on MMBC, the benchmark with the highest image reuse rate in Section~3.3. Accuracy rises as more tool-returned images remain available to later tools, increasing at each tested window size from 7.59\% with reuse disabled to 12.50\% with the full bank. Together with the reuse ablation in Section~3.3, this trend supports keeping images accessible throughout a search so that later tools can revisit earlier visual evidence.

\begin{figure}[t]
    \centering
    \includegraphics[width=\linewidth]{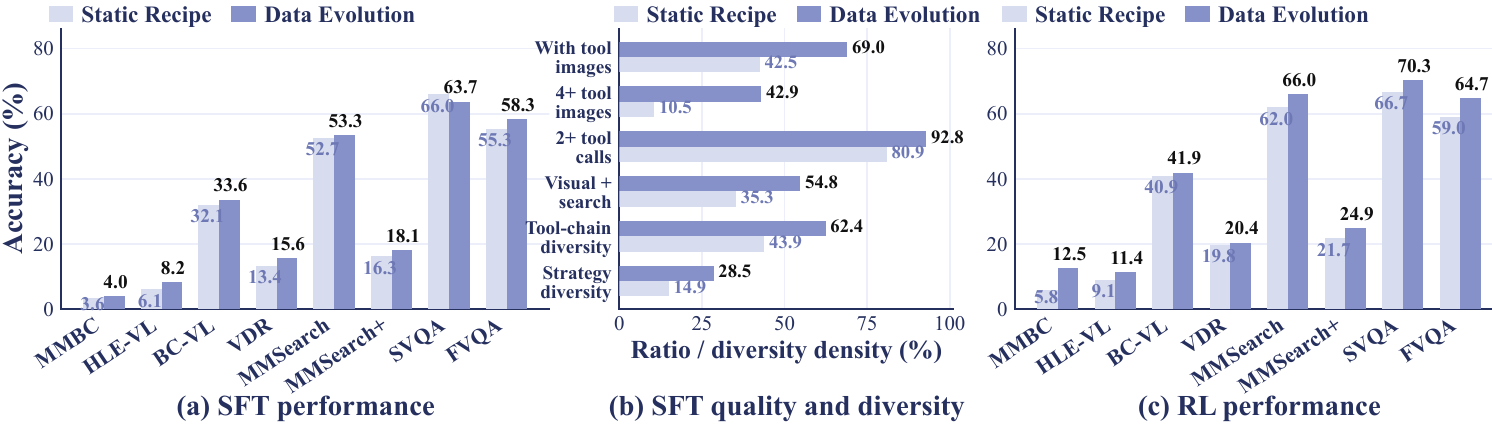}
    \caption{Static synthesis versus data evolution for agents based on Qwen3-VL-8B-Instruct.
    }
    \label{fig:static_vs_evolution}
\end{figure}

\subsection{Data Evolution vs. Static Synthesis}
To verify whether the gains of ODE come from closed-loop data evolution rather than from scaling a
fixed synthesis pipeline, we compare ODE with a static synthesis baseline that uses the initial ODE configuration and runs only the forward generation pipeline, without rollout-based analysis or configuration optimization. For SFT, we sample 2K traces from each source.
For RL, we start from the same ODE-8B-SFT checkpoint and train with two 4K RL
datasets, one produced by the evolved configuration and one produced by the static initial configuration.
Fig.~\ref{fig:static_vs_evolution} reports both downstream performance and the SFT trace statistics.

\textbf{Evolution improves downstream SFT performance.}
Fig.~\ref{fig:static_vs_evolution}(a) shows that SFT on evolved data outperforms the static recipe on most
benchmarks, with clear gains on visually grounded and search-oriented evaluations such as HLE-VL, VDR, MMSearch+, and FVQA. This indicates that the forward pipeline alone can generate usable teacher traces, but feedback-driven evolution produces more effective imitation data under the same sample scale. The result supports the central role of ODE: the benefit is not merely from generating synthetic data, but from adapting the synthesis configuration using rollout feedback.

\textbf{Evolved traces have higher quality and diversity.}
  Fig.~\ref{fig:static_vs_evolution}(b) explains why the evolved SFT data is more useful by comparing trace-level supervision patterns. \emph{With tool images} measures the fraction of traces containing at least one
  intermediate tool-produced image beyond the original task image, while \emph{4+ tool images} measures high-density visual supervision. \emph{2+ tool calls} captures multi-step tool use, and \emph{visual+search}
  measures whether a trace combines visual operations with search or browsing. Finally, tool-chain diversity
  counts distinct raw tool-call sequences, while strategy diversity groups these sequences into higher-level
  solving families.
  Compared with the static recipe, evolved traces contain more intermediate tool-produced images, a much larger
  fraction of high-density visual traces, more multi-step tool use, and more visual-search mixed strategies.
  The evolved subset also covers more distinct tool chains and broader abstract strategy families. Thus, ODE
  does not simply produce harder questions; it shifts the supervision distribution toward trajectories that
  demonstrate how to inspect visual evidence, combine tools, and solve tasks through richer agentic behavior.

\textbf{Data evolution improves RL performance.}
Fig.~\ref{fig:static_vs_evolution}(c) shows an even clearer pattern for RL. Starting from the same SFT checkpoint, RL on evolved data improves performance across the evaluated benchmarks, whereas RL on static data is weaker. This suggests that RL training benefits from tasks matched to the target policy's capabilities: a fixed initial recipe can generate tasks that are verifiable but poorly matched to the current policy's learning needs. ODE addresses this by using rollout feedback to move the generator toward tasks near the policy's learning frontier, making the resulting RL data more effective than static synthesis under the same data budget.

\begin{figure}[t]
    \centering
    \includegraphics[width=\linewidth]{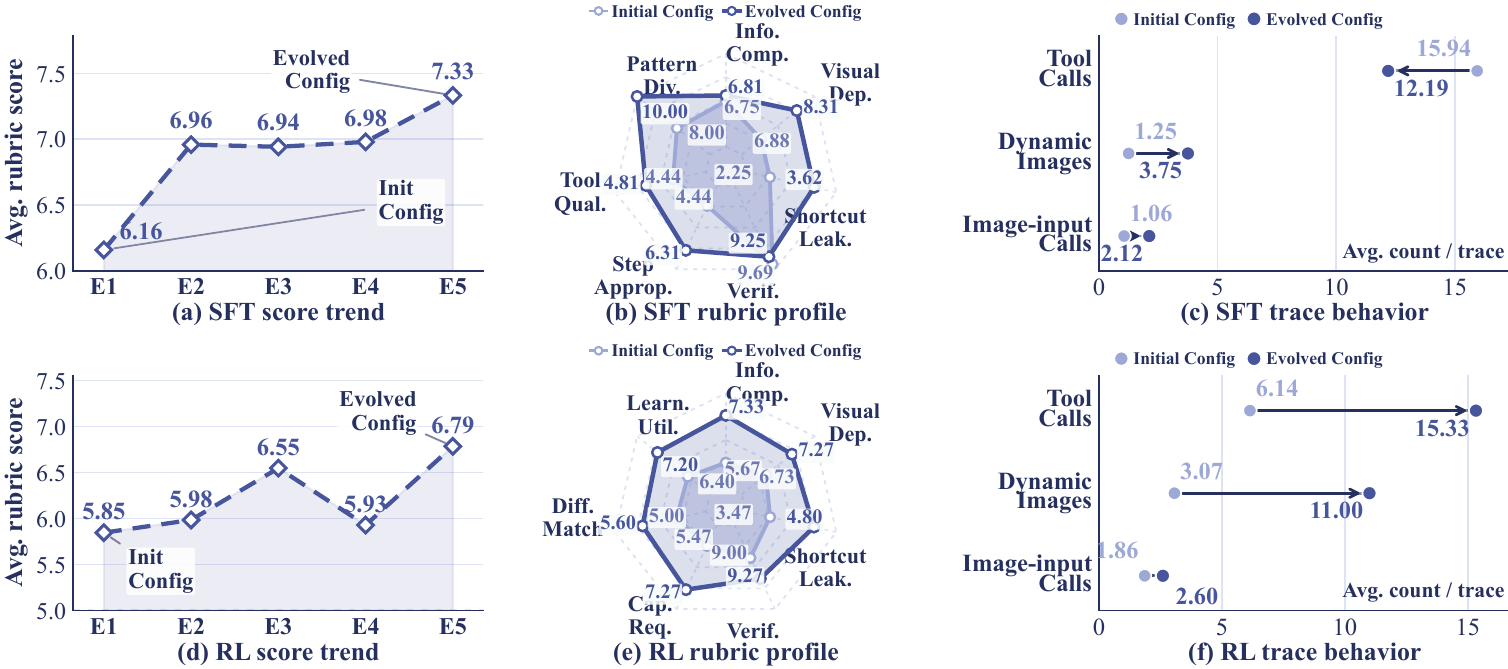}
    \caption{Mechanism analysis of ODE in SFT and 8B RL modes. Rubric scores are shifted by +5.}
    \label{fig:evolution_analysis}
\end{figure}

\subsection{Mechanism Analysis of ODE}
\label{sec:ode_mechanism}

  We further analyze what ODE changes during data construction. The goal is not only to show that evolved data
  performs better, but to understand how the closed-loop generator moves away from its initial configuration.
  We compare the initial and evolved configurations in both SFT and 8B RL modes.
  Fig.~\ref{fig:evolution_analysis} reports rubric-score trends, rubric profiles, and rollout-level behavior
  statistics. E1--E5 denote evolution rounds. Here, \emph{dynamic images} count tool-produced images acquired during rollout, excluding the
  original task image, and \emph{image-input calls} count tool invocations that take an image reference as
  input. Agreement with independent trace analyzers is reported in Appendix~\ref{app:analyzer_agreement}.

  \textbf{Evolution is mode-specific.}
  The same ODE loop produces different changes for SFT and RL, matching their different data objectives. In SFT
  mode, the evolved configuration mainly improves imitation-oriented dimensions such as visual dependency, step
  appropriateness, and tool-pattern diversity, while keeping verifiability high. This suggests that ODE does
  not simply make demonstrations longer or harder; it makes them better teacher traces. In RL mode, the
  improvements concentrate more on information complexity, capability requirement, difficulty match, and
  learning utility, indicating that the generator shifts toward tasks that are more suitable for policy
  improvement.

  \textbf{SFT traces become visually denser.}
  The behavior statistics show that evolved SFT data uses fewer tool calls overall, but introduces more dynamic
  images and more image-input calls. This is an important distinction: the evolved demonstrations are not
  better because they are longer. They are better because more of the supervision is carried by intermediate
  visual evidence, and later tool calls are more likely to operate on those visual states. This matches the
  intended role of ODE for SFT: selecting trajectories that teach the model how to inspect, reuse, and
  integrate visual evidence rather than merely execute many tools.

  \textbf{RL tasks induce deeper search.}
  For RL, evolution has a different behavioral effect. The evolved configuration induces rollouts with
  substantially more tool calls, more dynamic images, and more image-input calls. This indicates that ODE
  pushes the RL training task distribution toward examples that require active evidence gathering, rather
  than tasks solvable from the initial image or a single retrieval step. Together with the rubric improvements,
  this supports the central mechanism of ODE: rollout feedback steers the generator toward data that exposes
  the current policy's missing capabilities and provides a more useful training signal.

\section{Related Work}
Recent multimodal deep search agents extend MLLMs from static visual reasoning to interactive
evidence gathering over web and tool environments. Early systems such as MMSearch~\citep{jiang2025mmsearch} and Vision
Search Assistant~\citep{zhang2024visionsearchassistant} equip MLLMs with
image and web search, while newer benchmarks and agents emphasize harder settings that require
fine-grained visual grounding, iterative retrieval, and provenance verification~\citep{li2025mmbrowsecompcomprehensivebenchmarkmultimodal,tao2026mmsearchplus,geng2026webwatcher,huang2026visiondeepresearchincentivizingdeepresearchcapability}. Recent training-oriented systems
further improve multimodal search behavior with synthetic or semi-automated data and reinforcement
learning. MMSearch-R1~\citep{wu2025mmsearchr1incentivizinglmmssearch} curates multimodal
search VQA data for adaptive search, DeepMMSearch-R1~\citep{narayan2025deepmmsearchr1empoweringmultimodalllms}
builds web-grounded search data with crop-conditioned image search, WebWatcher~\citep{geng2026webwatcher}
combines synthetic cold-start trajectories with RL, and Vision-DeepResearch~\citep{huang2026visiondeepresearchincentivizingdeepresearchcapability} synthesizes long-horizon multi-tool
trajectories under retrieval noise. 
Our approach differs from these systems in how it connects visual tool use
with training data synthesis. In our visual-native harness, each tool-produced
image has a persistent reference that later tools can use to retrieve it.
ODE analyzes the resulting rollout traces and uses this feedback to adapt
the synthesis recipe to the agent's current learning needs.
See Appendix~\ref{app:related} for further discussion.

\section{Conclusion}

This paper presented a visual-native framework for multimodal deep search agents, combining a persistent image bank harness with On-policy Data Evolution (ODE). 
Across eight benchmarks, ODE-curated data consistently improves Qwen3-VL agents, increasing average accuracy from $24.9\%$ to $39.0\%$ at 8B and from $30.6\%$ to $41.5\%$ at 30B after SFT and RL. Further analyses show that reusable tool-generated images aid multi-step visual evidence gathering, while evolved data yields higher-quality and more diverse teacher traces than static synthesis. These findings highlight multimodal deep search as a co-design problem spanning the workspace, data generator, and training policy, and point to larger-scale on-policy evolution as a promising direction.

\renewcommand{\acksection}{\section*{Acknowledgements}}
\begin{ack}
This research was financially supported by the Open Research Fund from Guangdong Laboratory of Artificial Intelligence and Digital Economy (SZ), under Grant No. GML-KF-26-06. This work was also partially supported by the National Natural Science Foundation of China No. 625B2178. The authors declare no competing interests.
\end{ack}

\newpage

\bibliographystyle{plainnat}
\bibliography{refs}

\newpage

\appendix

\titlecontents{subsection}[4.2em]
  {}
  {\contentslabel{2.4em}}
  {\hspace*{-2.4em}}
  {\titlerule*[0.6pc]{.}\contentspage}

\section*{Appendix}
\addcontentsline{toc}{section}{Appendix}
\startcontents[appendix]
\printcontents[appendix]{l}{1}{\setcounter{tocdepth}{2}}
\clearpage

\section{Implementation Details of On-policy Data Evolution}
\label{app:ode_details}

Algorithm~\ref{alg:ode-preview} summarizes the On-policy Data Evolution loop described in Section~\ref{sec:method}. This appendix traces two consecutive rounds from a real run. We first lay out the round's frozen System Config and starting Evolvable Config $\mathcal{C}_t$ (Appendix~\ref{app:case_config}), and the seven-dimension trace rubric the analyzer scores against (Appendix~\ref{app:trace_rubrics}). We then follow a single seed through the four forward stages of round $t$, namely seed proposal, web exploration, graph organization, and task curation (Appendices~\ref{app:case_seed}--\ref{app:case_task}), with the actual images decoded back from the trace. We rollout the curated task with the policy under training, score it with the rubric, and read off the per-stage diagnoses, which the optimizer aggregates into four targeted edits that produce $\mathcal{C}_{t+1}$ (Appendix~\ref{app:case_update}). To show the loop actually closes, we then run round $t{+}1$ on $\mathcal{C}_{t+1}$ with a new seed and walk through its rubric scoring and the resulting update to $\mathcal{C}_{t+2}$ (Appendix~\ref{app:case_next_backward}), where the optimizer rolls back one of its round-$t$ edits in response to the new failure mode, showing that ODE's edits are responses to the policy's current weak point rather than monotone refinements along a fixed direction. 

\begin{algorithm}[htbp]
\captionsetup{labelsep=colon}
\caption{On-policy Data Evolution}
\label{alg:ode-preview}
\begin{algorithmic}[1]
\Require Mode $M$, rubric $R_M$, config $\mathcal{C}_0$, rounds $T$, fixed System Config and $\pi_{\mathrm{ref}}$
\Statex \hspace{\algorithmicindent}$\pi_{\mathrm{ref}}$: teacher for SFT; pre-RL target checkpoint for RL
\For{$t=0,\ldots,T-1$}
    \State $S_t \gets \textsc{SeedProposal}(\mathcal{C}_t)$
    \State $A_t \gets \textsc{WebExploration}(S_t,\mathcal{C}_t)$
    \State $G_t \gets \textsc{GraphOrganization}(A_t,\mathcal{C}_t)$
    \State $\mathcal{D}^{\mathrm{cand}}_t \gets \textsc{TaskCuration}(G_t,\mathcal{C}_t)$
    \For{each $x_i \in \mathcal{D}^{\mathrm{cand}}_t$}
        \State $\tau_i \gets \textsc{Rollout}(\pi_{\mathrm{ref}},q_i,\mathcal{I}_{0,i})$
        \State $\delta_i \gets \textsc{TraceAnalysis}(\tau_i,x_i,R_M)$
    \EndFor
    \State $\Delta_t \gets \textsc{Aggregate}(\{\delta_i\})$
    \State $P_t \gets \textsc{ProposeEdits}(\mathcal{C}_t,\Delta_t)$
    \State $\mathcal{C}_{t+1} \gets \textsc{AuditUpdate}(\mathcal{C}_t,P_t,\Delta_t)$
\EndFor
\State Select and freeze best audited config $\mathcal{C}^{*}$
\State $\mathcal{D}_{\mathrm{scale}} \gets \textsc{ForwardGenerate}(\mathcal{C}^{*})$
\If{$M=\mathrm{SFT}$}
    \State $\mathcal{U}_{\mathrm{scale}} \gets \textsc{Inputs}(\mathcal{D}_{\mathrm{scale}})$
    \State $\boldsymbol{\tau}_{\mathrm{scale}} \gets \textsc{Rollout}(\pi_{\mathrm{ref}},\mathcal{U}_{\mathrm{scale}})$
    \State $\mathcal{D}_{\mathrm{train}} \gets \textsc{RetainSFT}(\mathcal{D}_{\mathrm{scale}},\boldsymbol{\tau}_{\mathrm{scale}})$
    \State $\pi_\theta \gets \textsc{SFTTrain}(\mathcal{D}_{\mathrm{train}})$
\Else
    \State $\pi_\theta \gets \textsc{RLTrain}(\mathcal{D}_{\mathrm{scale}})$
\EndIf
\end{algorithmic}
\end{algorithm}

\definecolor{adeblue}{RGB}{35,82,124}
\definecolor{adeteal}{RGB}{24,105,96}
\definecolor{adeorange}{RGB}{164,91,22}
\definecolor{adepurple}{RGB}{91,65,150}
\definecolor{adeviolet}{RGB}{117,59,132}
\definecolor{adered}{RGB}{154,61,61}
\definecolor{adegreen}{RGB}{42,114,73}

\newtcolorbox{adebox}[2][]{
    enhanced,
    breakable,
    colback=adeblue!2!white,
    colframe=adeblue!85!black,
    coltitle=white,
    title=\textbf{#2},
    fonttitle=\small\bfseries,
    fontupper=\footnotesize,
    boxrule=0.7pt,
    arc=1.2mm,
    left=1.2mm,
    right=1.2mm,
    top=1.0mm,
    bottom=1.0mm,
    #1
}

\newtcolorbox{adecodebox}[2][]{
    enhanced,
    breakable,
    colback=black!2!white,
    colframe=black!55!white,
    coltitle=white,
    title=\textbf{#2},
    fonttitle=\small\bfseries,
    fontupper=\scriptsize\ttfamily,
    boxrule=0.65pt,
    arc=1.0mm,
    left=1.2mm,
    right=1.2mm,
    top=1.0mm,
    bottom=1.0mm,
    #1
}

\subsection{Round Configuration $\mathcal{C}_t$}
\label{app:case_config}

A round begins with two configuration objects, the System Config and the Evolvable Config $\mathcal{C}_t$. The System Config bundles every component that ODE deliberately freezes for the duration of an evolution run, so that backward refinement is comparing rounds under matched conditions. The Evolvable Config holds the four-stage generator parameters that the optimizer is allowed to edit between rounds. We give a sample of each below.

\begin{adecodebox}[colframe=adered!85!black]{System Config (frozen for the run)}
mode                 : rl\\
rollout\_model        : the policy currently under training, queried through the visual-native harness\\
judge                : LLM judge that compares the rollout's final answer to the reference and emits a success or failure label\\
analyzer             : LLM analyzer that scores each rollout against the round's rubric and returns the per-rollout diagnosis $\delta_i$\\
rubric               : RL rubric, $7$ dimensions, weights {[}1.0, 1.2, 1.2, 1.0, 1.0, 2.0, 1.6{]}\\
tool\_environment     : nine harness tools (web\_search, web\_fetch, image\_search, visual\_search, link\_reader, zoom\_in, rotation, flip, python\_code)\\
seed\_type            : entity\_with\_image\\
sampling\_axes        : 11 domains, 4 ability profiles, 4 difficulty levels
\end{adecodebox}

\begin{adecodebox}[colframe=adeblue!85!black]{$\mathcal{C}_t$, the round's Evolvable Config (key numerical fields)}
seed\_proposer.max\_steps         = 8\\
explorer.max\_steps               = 10\\
explorer.params.number\_of\_anchors = 6\\
explorer.params.max\_nodes\_per\_phase = 2\\
explorer.params.image\_ratio      = 0.5\\
graph\_organizer.complexity.reasoning\_ratio       = 0.33\\
graph\_organizer.complexity.perception\_ratio      = 0.40\\
graph\_organizer.complexity.reasoning\_max\_steps  = 5\\
graph\_organizer.complexity.perception\_max\_steps = 4\\
curator.few\_shot\_difficulty\_weights = \{easy: 0.05, medium: 0.15, hard: 0.50, expert: 0.30\}
\end{adecodebox}

The key prompt fields used by the loop are embedded in the corresponding stage cases below. The forward-stage prompts belong to $\mathcal{C}_t$ and evolve with the numerical fields: the optimizer can append rejection rules, swap clauses, or rephrase strategy hints based on the stage diagnoses returned by backward refinement.

\subsection{Mode-Specific Trace Rubrics}
\label{app:trace_rubrics}

The analyzer scores each rollout along a seven-dimension rubric. Every dimension takes an ordinal score from $-5$ to $+5$ with a short textual justification, and the overall score $s(\delta_i)$ is the weighted average of the seven dimension scores using the weights listed below. The analyzer also returns a stage-level attribution field that names which of \texttt{seed\_proposer}, \texttt{explorer}, \texttt{graph\_organizer}, or \texttt{curator} is responsible for any observed failure. Four dimensions, namely \emph{Information\_Complexity}, \emph{Visual\_Dependency}, \emph{Shortcut\_Leakage}, and \emph{Verifiability}, are shared across modes because they describe the task itself regardless of how it is consumed. The remaining three dimensions are mode-specific.

\begin{adebox}[colframe=adepurple!85!black,colback=adepurple!2!white]{SFT Rubric: Teacher Trace Value}
The SFT rubric favours teacher trajectories that are useful for distillation, rather than merely correct final answers.
\textbf{Information\_Complexity} (weight $1.0$). Rewards traces teaching multi-source search, cross-reference, and synthesis. Rejects common-knowledge answers and one-query lookups.
\textbf{Visual\_Dependency} (weight $1.2$). Rewards traces where visual interpretation is necessary and visible in the reasoning process. Rejects text-only tasks and irrelevant visual context.
\textbf{Shortcut\_Leakage} (weight $1.5$). Rewards traces whose key cues must be actively discovered. Rejects examples where the answer is exposed by the question, an obvious image cue, or the first search snippet.
\textbf{Verifiability} (weight $1.0$). Rewards objective answers with a clear evidence-to-answer chain. Rejects subjective, ambiguous, or weakly judgeable conclusions.
\textbf{Step\_Appropriateness} (weight $1.2$). Rewards traces whose number and type of steps match the task difficulty and ability profile. Rejects demonstrations that teach miscalibrated effort.
\textbf{Tool\_Usage\_Quality} (weight $1.5$). Rewards purposeful tool selection, chaining, verification, and stopping behaviour. Rejects circular calls, redundant tools, or missing critical operations.
\textbf{Tool\_Pattern\_Diversity} (weight $3.0$). A batch-level metric computed from complete tool-call sequences, encouraging diverse strategy types such as search-only, visual-search, computation-assisted, and multi-step visual workflows.
\end{adebox}

\begin{adebox}[colframe=adeviolet!85!black,colback=adeviolet!2!white]{RL Rubric: Task Quality and Policy Suitability}
The RL rubric favours tasks that are judgeable, evidence-supported, and useful for optimizing the current policy.
\textbf{Information\_Complexity} (weight $1.0$). Rewards tasks requiring multi-source evidence integration. Rejects tasks answerable by memorization, common knowledge, or a single obvious query.
\textbf{Visual\_Dependency} (weight $1.2$). Rewards tasks whose solving process depends on visual information from the input or tool-returned images. Rejects text-only solutions or misleading visual context.
\textbf{Shortcut\_Leakage} (weight $1.2$). Rewards tasks whose key cues require effort to uncover and verify. Rejects tasks where the answer is leaked by the prompt, obvious image content, or search snippets.
\textbf{Verifiability} (weight $1.0$). Rewards tasks with unique, objective answers verifiable against authoritative evidence. Rejects subjective or under-specified tasks.
\textbf{Capability\_Requirement} (weight $1.0$). Rewards tasks requiring multi-step reasoning, tool planning, and integration of visual and textual evidence. Rejects tasks solvable without meaningful tools or reasoning.
\textbf{Difficulty\_Match} (weight $2.0$). Rewards tasks near the policy's learning zone. The analyzer maps each rollout to a five-stage difficulty diagnosis, ranging from tasks already solved with little effort, through under-challenging tasks and productive matches, to tasks beyond the policy's current capability or made hard by brittle evidence and infrastructure noise. The optimizer reads this distribution when steering batch-level difficulty.
\textbf{Learning\_Utility} (weight $1.6$). Rewards tasks and traces with clear correctness signal, attributable errors, transferable strategies, and useful capability gaps. Rejects chaotic, misleading, or non-transferable traces.
\end{adebox}

\subsection{Stage 1: Seed Proposal}
\label{app:case_seed}

\begin{figure}[h]
    \centering
    \includegraphics[height=4.4cm]{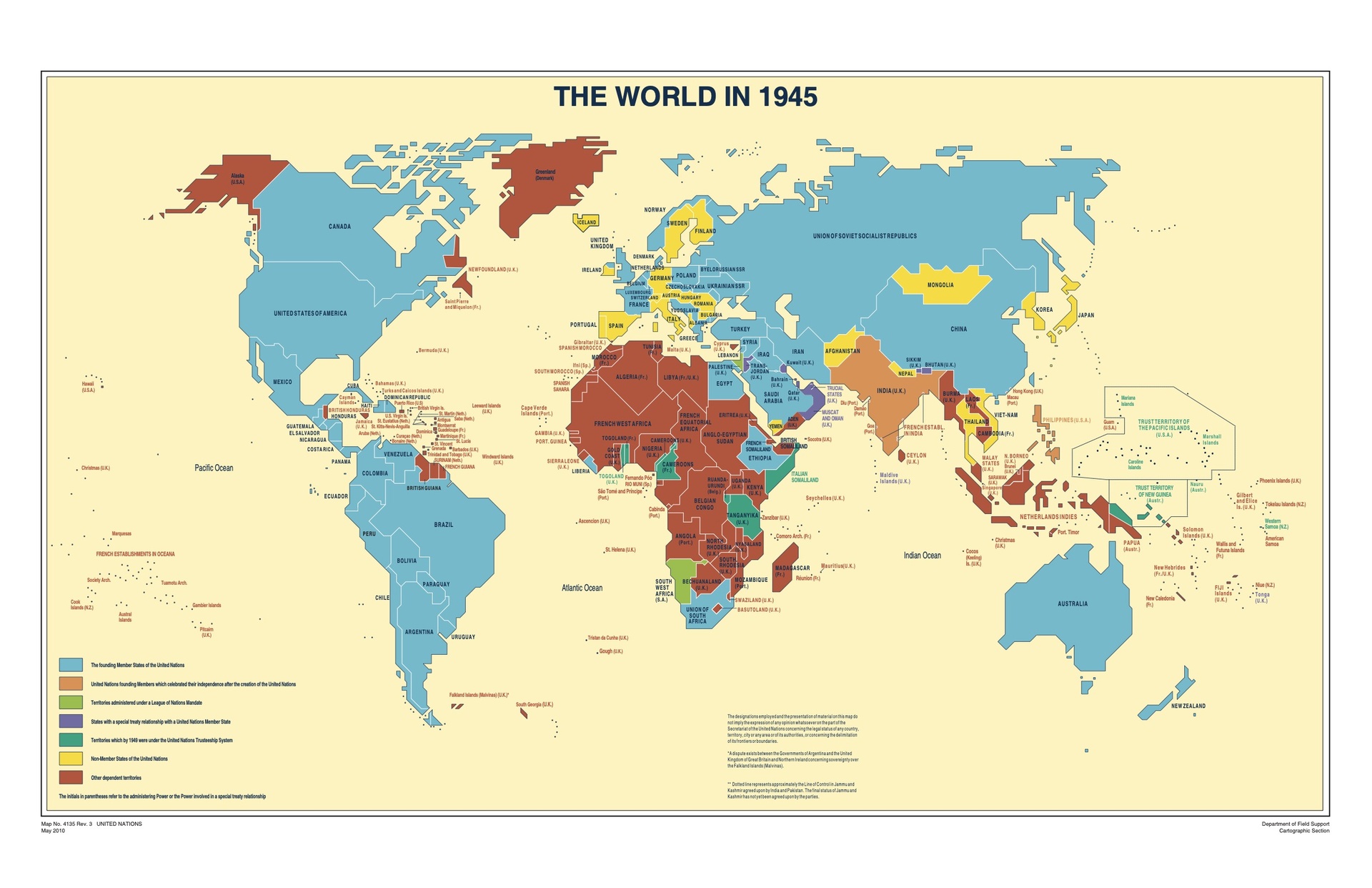}
    \caption{\textbf{Seed image $\mathcal{I}_0$.} The seed proposer samples an \texttt{entity\_with\_image} pair anchored on \emph{United Nations Map No.~4135 Rev.~3, ``The World in 1945''} (May 2010), domain \texttt{geography}.}
    \label{fig:case_seed}
\end{figure}

\begin{adebox}[colframe=adeteal!85!black,colback=adeteal!2!white]{Seed Record}
\textbf{Entity.} United Nations Map No.~4135 Rev.~3: The World in 1945 (May 2010).
\textbf{Domain.} \texttt{geography}.
\textbf{Visual potential.} The map carries legible, visually extractable details, including the official numeric map identifier \texttt{4135 Rev.~3}, publication date \texttt{May 2010}, and the thematic legend categories for 1945 global territorial statuses, all of which require direct visual inspection rather than keyword recognition.
\textbf{Information depth.} The seed supports multi-hop reasoning, such as cross-referencing the map number and revision against the UN cartographic archive, or validating the 1945 boundary classifications against independent historical geographic datasets, and rejects trivial single-snippet solutions.
\textbf{Search queries.} \texttt{United Nations map "Map No." "Rev." jpg scale bar}; \texttt{3 May 2010 United Nations cartographic section world in 1945}.
\textbf{Source URLs.} Cornell University Press \emph{UN Support for Decolonization} page; UN file repository at \texttt{un.org/id/file/59483}.
\textbf{Judge decision.} Accepted. The judge confirms that the artifact carries information-bearing image content and supports multi-hop verifiable tasks.
\end{adebox}

\begin{adebox}[colframe=adeteal!85!black,colback=adeteal!2!white]{Seed Proposal Prompt Fields}
\textbf{\texttt{strategy}.} The seed proposer is instructed to find \texttt{entity\_with\_image} seeds for deep search tasks. A good seed must pair a real entity with an information-bearing image and enough factual surface area to support multi-hop verification and computation. The prompt prioritizes labeled maps, museum placards, technical diagrams, archival documents, posters, charts, timelines, and other images whose labels, dates, quantities, coordinates, or legends can be visually extracted. It also requires domain diversity across the seed batch and rejects decorative photos, common-knowledge trivia entities, paywalled images, and unstable social posts.
\textbf{\texttt{default\_requirement}.} The acceptance checklist requires \texttt{web\_search}, \texttt{image\_search}, and at least one \texttt{visual\_search} call. The image must be relevant, high-resolution, and readable under zoom, and the entity or visible facts must be supported by at least two independent sources, preferably authoritative institutions such as museums, universities, governments, publishers, or recognized organizations. The seed must also support multi-hop lookup plus at least one reasoning operation, rather than a single-snippet answer.
\textbf{\texttt{seed\_prompt}.} The agent is asked to propose a small set of diverse multimodal seeds. For each candidate it searches for a factual entity, locates an information-bearing image, verifies that the image contains readable content, and finds an independent corroborating source. The required output is a YAML-style record containing \texttt{entity}, \texttt{entity\_type}, \texttt{image\_url}, \texttt{image\_source\_page}, \texttt{supporting\_sources}, \texttt{why\_visual}, \texttt{multi\_hop\_potential}, and \texttt{rejection\_risks}.
\end{adebox}

\subsection{Stage 2: Web Exploration}
\label{app:case_exploration}

The explorer expands the seed into a small information network of six anchors, visiting twelve URLs over a single exploration phase. Anchors include the cartographic baseline, the September 1948 UN snapshot, the institutional pathway from the Trusteeship Council to the C-24 Special Committee on Decolonization, the modern Non-Self-Governing-Territories list, and territory-specific maps. Each anchor records textual facts, source URLs, and tool-returned image identifiers. Two of the anchor images are reproduced below.

\begin{figure}[h]
    \centering
    \begin{subfigure}[t]{0.46\linewidth}
        \centering
        \includegraphics[height=3.6cm]{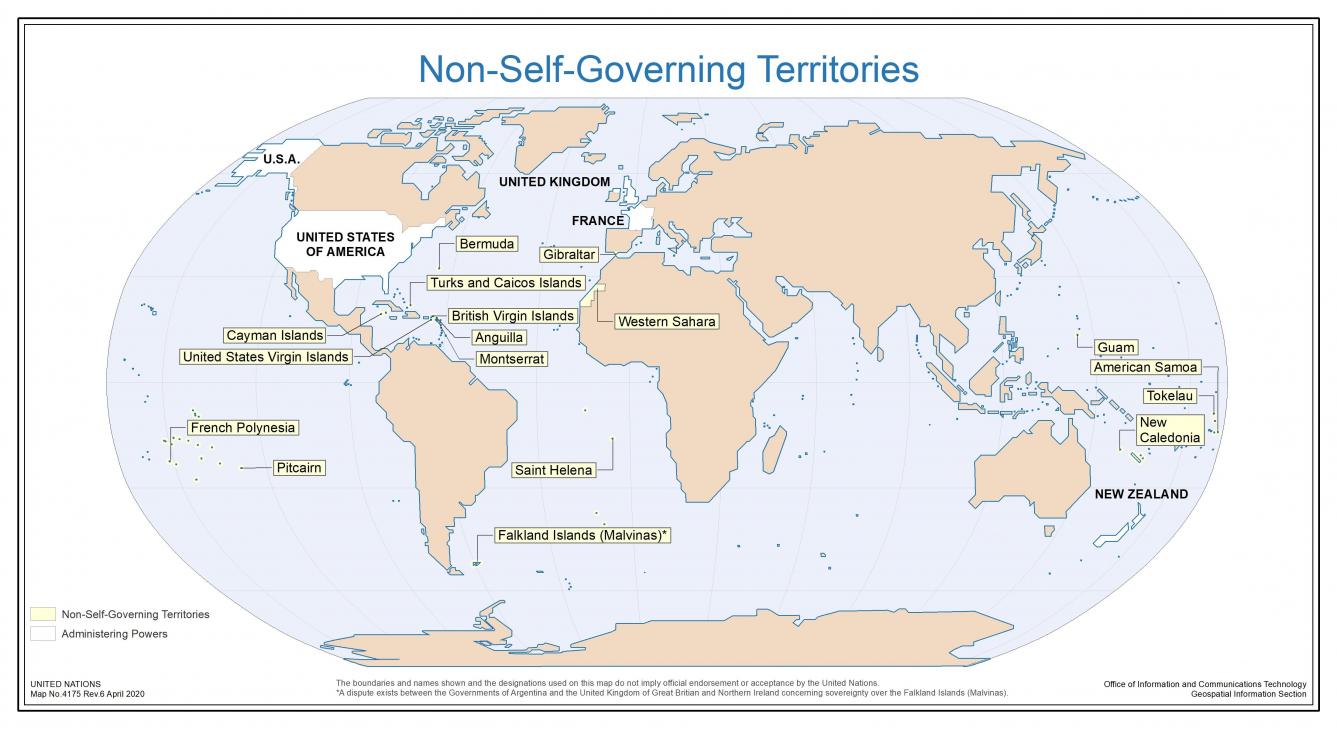}
        \caption{The contemporary NSGT global map, \emph{UN Map No.~4175 Rev.~6 (April 2020)}.}
        \label{fig:case_expl_a}
    \end{subfigure}\hfill
    \begin{subfigure}[t]{0.46\linewidth}
        \centering
        \includegraphics[height=3.6cm]{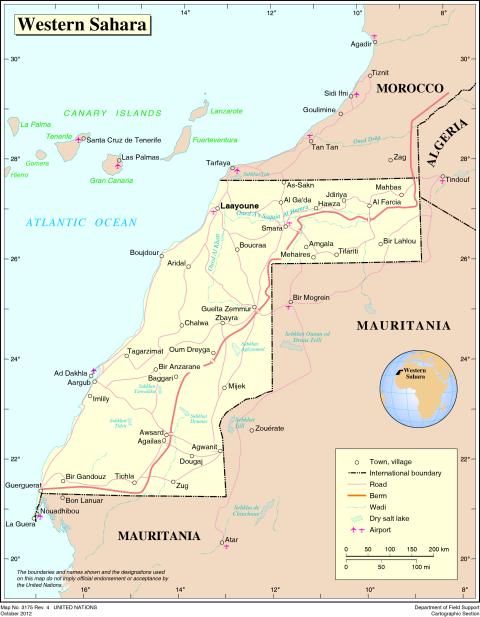}
        \caption{The territory-specific UN reference map for Western Sahara, \emph{UN Map No.~3175 Rev.~5 (Jan 2020)}.}
        \label{fig:case_expl_b}
    \end{subfigure}
    \caption{\textbf{Tool-returned anchor images} from the explorer. Each is appended to the image bank under a fresh \texttt{\string<image: N\string>} identifier and remains available to later stages and to the rollout policy.}
    \label{fig:case_expl}
\end{figure}

\begin{adebox}[colframe=adeorange!85!black,colback=adeorange!3!white]{Explorer Record}
\textbf{Topic.} UN cartography of post-WWII territorial status.
\textbf{Visited URLs.} 12 (UN Geospatial Information Section, UN Charter texts, Trusteeship Council documents, NSGT roster, Western Sahara reference page, Britannica, and Wikipedia event pages).
\textbf{Anchors collected.} (1)~UN Map No.~4135 Rev.~3, baseline 1945 world; (2)~UN Map of UN Members and Dependent Territories, September 1948; (3)~UN Trusteeship Council; (4)~Special Committee on Decolonization (C-24); (5)~UN Map No.~4175 Rev.~6, NSGT global map (April 2020); (6)~UN Map No.~3175 Rev.~5, Western Sahara (January 2020).
\textbf{Quantitative facts attached.} The 11 ``original'' UN Trust Territories administered under Chapter~XII; the September 1948 snapshot showing 10 Trust Territories already administered (Italian Somaliland's trusteeship begins only in 1950); 17 NSGTs listed in the 2020 NSGT roster.
\end{adebox}

\begin{adebox}[colframe=adeorange!85!black,colback=adeorange!3!white]{Web Exploration Prompt Fields}
\textbf{\texttt{strategy}.} The explorer is instructed to expand the seed into a compact knowledge graph for multimodal deep search QA generation. Each anchor must add a new direction, such as a related artifact or version, upstream cause, downstream impact, comparable peer entity, geographic context, dataset or report, or historical milestone. The prompt favors anchors with dates, measurements, counts, money, coordinates, legend structures, classification tables, competing claims, or cross-modal details that appear in images but not plainly in text.
\textbf{\texttt{quality\_requirements}.} Every node must be built from fresh tool use rather than memory. The prompt requires at least two distinct tool calls per node, a clear definition, filled detailed-explanation fields, six to ten key facts, at least three numeric or date facts when available, and at least two independent sources. For image-bearing nodes, the record must state exactly what can be extracted visually and why the image is useful downstream.
\textbf{\texttt{exploration\_process\_prompt}.} The loop starts from the seed or previous node, identifies a promising next anchor, uses \texttt{web\_search} to locate authoritative pages and alternatives, uses \texttt{link\_reader} to extract precise details, and uses \texttt{image\_search} plus \texttt{visual\_search} when the anchor should satisfy the image-evidence target. Each node is written with title, type, definition, detailed explanation, key facts, images, sources, and source-support notes before the explorer chooses the next anchor. The prompt explicitly prefers fewer high-quality facts over long vague summaries.
\end{adebox}

\subsection{Stage 3: Graph Organization}
\label{app:case_graph}

The graph organizer assembles the anchors into a multimodal evidence graph $G$ and enriches it with reasoning and perception nodes that test cross-source consistency and surface fine-grained visual details for the curator to anchor on.

\begin{adebox}[colframe=adepurple!85!black,colback=adepurple!2!white]{Organized Graph}
\textbf{Core themes.} (1)~UN cartographic documentation of territorial status, 1945 to 2020. (2)~The shift from colonial dependencies and trust territories to the NSGT framework. (3)~UN institutional oversight of decolonization, from the Trusteeship Council to the C-24 Special Committee. (4)~Territory-specific case documentation (Western Sahara).
\textbf{Edge structure.} Node~1 (1945 baseline map) provides the territorial-status framework that later UN maps extend. Node~2 (1948 UN snapshot) \emph{supports} Node~3 by depicting Trust Territories under the Trusteeship Council. Node~3 \emph{explains} the governance mechanism behind Node~2 and is functionally extended by Node~4 (C-24). Node~4 \emph{exemplifies} the modern NSGT pathway and is realized as the cartographic product Node~5. Node~6 (Western Sahara) \emph{supports} Node~5 through territory-specific cartographic detail.
\textbf{Reasoning enrichment.} A consistency check between (a)~the count of Trust Territories visible on the 1948 map and (b)~the canonical roster of 11 originals attaches a derived quantity ``$10/11 \approx 90.9\%$'' to the edge connecting Node~2 and Node~3, with provenance to the Trusteeship Council page. The check fires under \texttt{graph\_organizer.complexity.reasoning\_ratio} = 0.33 in $\mathcal{C}_t$.
\textbf{Perception enrichment.} A \texttt{zoom\_in} pass over the 1948 map legend exposes the shading category for Trust Territories and is added back as a new perception node, and a second \texttt{zoom\_in} over the Western Sahara map legend extracts the administering-power annotation. The pass fires under \texttt{graph\_organizer.complexity.perception\_ratio} = 0.40 in $\mathcal{C}_t$.
\end{adebox}

\begin{adebox}[colframe=adepurple!85!black,colback=adepurple!2!white]{Graph Organization Prompt Fields}
\textbf{\texttt{organization\_strategy}.} The organizer is asked to analyze the relationships among collected information nodes and turn them into a coherent knowledge graph. The prompt receives each node's anchor, node type, core information, connections, and available images, then asks for themes, edges, source relations, entity or event relations, and cross-modal dependencies.
\textbf{\texttt{complexity.reasoning\_strategies\_prompt}.} The reasoning agent is asked to create a reasoning-enhanced node through analysis or computation. It can call \texttt{python\_code} and \texttt{visit}, must use at least one tool, and should add analytical value beyond restating the original node. The output schema records the reasoning type, process, tools used, findings, computation when applicable, source URLs, and the connection back to the original anchor.
\textbf{\texttt{complexity.perception\_strategies\_prompt}.} The perception agent is asked to create a perception-enhanced node by operating on an image. It can call \texttt{zoom\_in}, \texttt{flip}, \texttt{rotation}, and \texttt{visual\_search}. The prompt emphasizes choosing transformations that reveal new information, such as small labels, legends, angled text, logos, species, or landmarks, and it requires the resulting visual finding to be documented as a new graph node rather than an informal note.
\textbf{\texttt{complexity.enhancement\_requirements}.} Both enhancement paths must produce nodes that are useful for later task curation. The enhanced node should have a clear definition, explanation, key facts, provenance, and explicit connections so that the curator can ground questions in the enriched graph.
\end{adebox}

\subsection{Stage 4: Task Curation}
\label{app:case_task}

\begin{figure}[h]
    \centering
    \includegraphics[height=4.6cm]{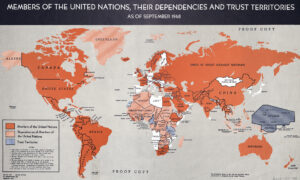}
    \caption{\textbf{Curated task image} for the worked example. The image is the September 1948 UN snapshot, selected from the evidence graph as the visual anchor of the curated question. It is registered into the image bank as $\mathcal{I}_0$ before rollout.}
    \label{fig:case_task}
\end{figure}

\begin{adebox}[colframe=adeviolet!85!black,colback=adeviolet!2!white]{Curated Task Record}
\textbf{Question.} ``In the world map shown (dated `as of September 1948'), use the legend to identify which shaded regions count as UN-administered trust territories (not ordinary dependencies/colonies), and count how many distinct trust territories are actually depicted. Then compare that count to the full set of `original' trust territories administered under the UN's trusteeship system by the UN organ responsible for trust territories. Be careful to exclude the Horn-of-Africa territory that is often confused with similarly named `Somaliland' regions but whose trusteeship agreement started later (in 1950, so it would not appear as a trust territory on a September 1948 map). What percent of the original trust-territory set were already designated as trust territories by September 1948? Round to the nearest whole percent.''
\textbf{Ground truth.} \texttt{91\%}.
\textbf{Difficulty.} \texttt{hard} (sampled under $\mathcal{C}_t$.curator.few\_shot\_difficulty\_weights with weight 0.50).
\textbf{Ability profile.} \texttt{perception\_reasoning}.
\textbf{Visual dependency.} The map legend is required to distinguish trust territories from other dependencies, and the small on-map labels are required to count the distinct trust territories depicted. The count cannot be obtained reliably without reading the map.
\textbf{Evidence chain.} The 1948 UN map's trust-territory legend category corresponds to 10 distinct labeled trust territories. The UN Trusteeship Council administered 11 original trust territories in total. The missing one relative to the originals is the former Italian-administered Somali territory whose trusteeship began only in 1950 and therefore does not appear on a September 1948 map, so the share is $10/11 \approx 90.9\%$, rounding to 91\%.
\textbf{Planned reasoning steps.} Nine steps mixing image\_processing (read effective date, identify legend category, scan and label trust territories), web\_search (retrieve the original-set count and the Somaliland exclusion), and calculate (form the percentage and round). Curator complexity-enhancement rewrote an earlier easier draft (``what percentage of original trust territories were already designated by 1948?'') into the present form by introducing the explicit Somaliland-exclusion check and removing the verbatim total.
\end{adebox}

\begin{adebox}[colframe=adeviolet!85!black,colback=adeviolet!2!white]{Task Curation Prompt Fields}
\textbf{\texttt{difficulty\_control\_prompt}.} The curator is instructed to avoid trivial tasks, avoid brittle or unsolvable tasks, and prefer hard-but-learnable examples. The desired task should combine at least one key image fact, one web-only fact, and one synthesis, comparison, or computation step. Expert difficulty is encouraged only when the evidence path remains clean and judgeable.
\textbf{\texttt{strategy}.} The task-generation prompt turns the evidence graph into diverse QA tasks. It requires short, verifiable answers such as entities, numbers, dates, or names, and it discourages sentence-length ground truth. It also enforces multi-step tool use, specialized knowledge beyond common facts, diverse ability profiles, and question phrasing that asks what to find rather than how to find it.
\textbf{\texttt{quality\_requirements\_prompt}.} The final self-check covers answer format, answer uniqueness, objective verifiability, complete evidence chain, natural image references, absence of numbered image indices, no tool-name or strategy leakage, specialized knowledge complexity, and batch diversity. The prompt bans questions that say ``zoom in'', ``search for'', ``use OCR'', ``calculate by'', or otherwise reveal the solving procedure.
\textbf{\texttt{complexity\_enhancement.requirements\_prompt}.} The enhancement pass preserves the ground-truth answer while increasing reasoning depth. It keeps image references natural, avoids technical image tokens, removes strategy leakage, targets longer hard or expert reasoning chains, and ensures the enhanced question remains answerable from the graph evidence.
\textbf{\texttt{complexity\_enhancement.strategy\_prompt}.} The rewrite strategies include removing obvious clues, replacing direct entity names with relationship-based descriptions, adding intermediate lookups, adding comparisons or computations, converting single-hop questions into multi-hop questions, and preserving visual dependency when the task is multimodal.
\end{adebox}

\subsection{Stage 5: Rollout and Stage 6: Backward Analysis}
\label{app:case_analysis}

The candidate task is rolled out by the policy under training. The verifier finds the final answer incorrect, and the analyzer then scores the rollout along the rubric and attributes observed failures back to specific forward stages.

\begin{adecodebox}[colframe=adered!85!black]{Backward Analysis Prompt Fields}
rubric\_prompt: You are a professional data quality evaluator. Score the Agent execution trace based on the rubric. For each dimension, read the question, ground truth, visual materials, complete reasoning steps, tool calls, observations, and outcome, then return a score and explanation.\\
scoring\_principles: evaluate QA quality through task design and execution process. Synthesize evidence from the question, ground truth, visual materials, rollout steps, tool calls, observations, and final answer.\\
Difficulty\_Match rule: classify the rollout into the five-stage difficulty diagnosis from too simple to too difficult or noisy. Do not reward long traces, repeated retries, or external failures by themselves. High scores require productive struggle with concrete intermediate progress.\\
diagnosis\_prompt: after rubric scoring, diagnose root causes using the forward construction record. Attribute each issue to seed\_proposer, explorer, graph\_organizer, or curator, report severity, affected dimensions, and a suggested config-side repair.
\end{adecodebox}

\begin{adebox}[colframe=adered!85!black,colback=adered!2!white]{Rubric Scores (overall $3.82$, cognitive friction \texttt{optimal})}
\textbf{Information\_Complexity (4.0).} Multi-source synthesis of legend-based classification on the 1948 map plus the canonical 11-original Trusteeship set, with cross-checks for the 1950 Somaliland exclusion.
\textbf{Visual\_Dependency (5.0).} The depicted-count of trust territories on the September 1948 map is recoverable only by visual interpretation of shaded regions and labels. Text-only sources can give the original total (11) but not the map's depicted count.
\textbf{Shortcut\_Leakage (3.0).} The original-set total is one snippet away from any UN page, but the depicted count still requires careful map reading. The final percentage is not directly leaked.
\textbf{Verifiability (5.0).} Single objective numeric target, rounded to whole percent.
\textbf{Capability\_Requirement (5.0).} Demands precise zoom and crop to read the legend and small territory labels, correct entity classification (trust vs.\ mandate vs.\ dependency), external verification of the originals list, and arithmetic with rounding.
\textbf{Difficulty\_Match (3.0, tag {\small\texttt{[good\_match]}}).} Challenging but achievable. The agent failed, but on correctable issues (missed visual items, misclassification of South-West Africa as a trust territory), placing the task inside the policy's productive learning zone.
\textbf{Learning\_Utility (3.0).} Failure mode is clearly attributable to visual extraction quality and behaviour, namely prioritizing accurate legend reading over speculative web search and validating each shaded region against the legend before counting. The trace yields transferable training signal.
\end{adebox}

\begin{adebox}[colframe=adeblue!85!black,colback=adeblue!2!white]{Stage Diagnoses (analyzer output, abridged)}
\textbf{seed\_proposer (severity moderate).} The seed did not robustly lock identity and provenance to a single canonical map artifact (map number, revision, date), so the downstream pipeline could drift to a different UN-themed 1948 map than the proposed seed. The seed also did not enforce a high-resolution image requirement, which contributed to later misclassifications when low-resolution copies were used. \emph{Affected dimensions.} Difficulty\_Match, Learning\_Utility, Visual\_Dependency, Verifiability.

\textbf{explorer (severity severe).} The explorer collected only six anchors with thin per-anchor evidence and did not produce zoom-validated visual extractables for the legend or small territory labels. \emph{Affected dimensions.} Information\_Complexity, Visual\_Dependency, Capability\_Requirement.

\textbf{graph\_organizer (severity moderate).} The reasoning enrichment recorded the $10/11$ derivation but did not run a perception enrichment that would have surfaced the legend's shading categories as separate nodes. \emph{Affected dimensions.} Visual\_Dependency, Capability\_Requirement.

\textbf{curator (severity moderate).} The question is well-formed, but its evidence chain depends on a low-resolution count that the rollout policy plausibly miscounts. \emph{Affected dimensions.} Verifiability, Difficulty\_Match.
\end{adebox}

\subsection{Optimizer Update to $\mathcal{C}_{t+1}$}
\label{app:case_update}

The optimizer aggregates the per-rollout diagnoses across the round's batch into a single $\Delta_t$, applies the implied edits to the corresponding fields of $\mathcal{C}_t$, and writes out $\mathcal{C}_{t+1}$. The next-round config differs from $\mathcal{C}_t$ at exactly the four numerical fields below. The string-valued strategy and requirement prompts are updated in parallel by appending the analyzer's suggested rejection rules verbatim.

\begin{adecodebox}[colframe=adegreen!85!black]{Optimizer Prompt Fields}
role: You are a professional AI training data system architect. Based on rubric score analysis and diagnosis feedback, generate precise, structured changes to the stage configuration.\\
inputs: rubric score table, dimension trends, bottom-tail failure modes, batch health, policy pass rate, current stage config, available field paths, programmatic issue aggregation, stable fault signals, cross-stage root-cause signal, and stage-local rubric responsibility.\\
actions: update\_param for numerical or boolean parameters, append\_text for new prompt clauses, replace\_text for exact-substring edits, and rewrite\_text for consolidating an entire prompt field.\\
constraints: numerical edits are small, prompt edits must be surgical, patch sets should be compact, parser noise is ignored, current-config faults are fixed before weak-reference drift, schema contracts are preserved, and stable recurring failures are preferred over case-specific rules.\\
reviewer\_prompt: a conservative reviewer filters proposed changes, keeping only edits directly supported by recurring fault signals, compatible with the config contract, not case-specific overfitting, and not unnecessary prompt bloat.
\end{adecodebox}

\begin{adecodebox}[colframe=adegreen!85!black]{Diff $\mathcal{C}_t \to \mathcal{C}_{t+1}$ (numerical fields)}
seed\_proposer.max\_steps              : 8 $\to$ 10\\
explorer.params.image\_ratio          : 0.50 $\to$ 0.40\\
graph\_organizer.complexity.reasoning\_max\_steps  : 5 $\to$ 6\\
graph\_organizer.complexity.perception\_max\_steps : 4 $\to$ 5
\end{adecodebox}

The four edits map cleanly onto the four stage diagnoses above. (1)~Raising \texttt{seed\_proposer.max\_steps} from $8$ to $10$ gives the seed proposer enough budget to satisfy the new identity-and-provenance lock the optimizer appended to its \texttt{default\_requirement}, addressing the seed-stage drift. (2)~Lowering \texttt{explorer.params.image\_ratio} from $0.50$ to $0.40$ trades off raw image-fetch frequency for tighter visual-evidence quality, paired with a separate prompt edit that requires a \texttt{visual\_search} validation per quantitative anchor, addressing the explorer's thin per-anchor evidence. (3,4)~Raising both reasoning and perception \texttt{max\_steps} by one in the graph organizer gives complexity enhancement room to attach legend-category nodes and run an extra cross-source consistency check, addressing the missing perception enrichment. The curator field set is left unchanged, because the curator's diagnosis is a downstream consequence of the explorer and graph-organizer issues that the optimizer now addresses upstream. The batch-level pass-rate signal also leaves \texttt{curator.few\_shot\_difficulty\_weights} unchanged, since the {\small\texttt{[too\_hard]}} share is below the threshold that would trigger a difficulty-weight shift.

\subsection{Round $t{+}1$ Forward Under $\mathcal{C}_{t+1}$}
\label{app:case_next}

To show that the loop actually closes, we walk through the next round on $\mathcal{C}_{t+1}$. The same optimizer that produced $\mathcal{C}_{t+1}$ now drives a new seed through the four forward stages, and the rubric scores expose a different failure mode that the next update will then address.

\begin{figure}[h]
    \centering
    \begin{minipage}[t]{0.46\linewidth}
        \centering
        \includegraphics[height=4.0cm]{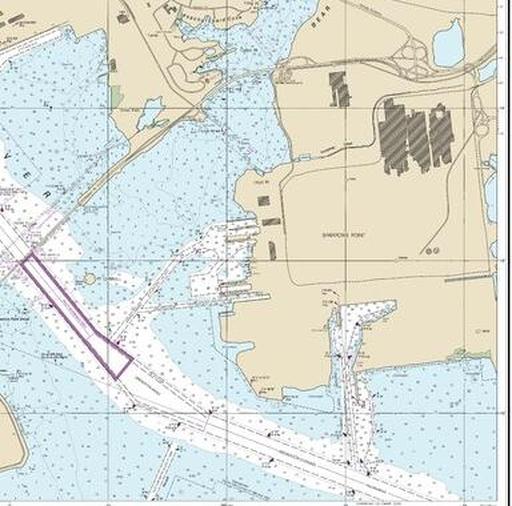}
        \caption*{\footnotesize\emph{Round $t{+}1$ exploration anchor 4}: zoomed segment of the Seagirt Marine Terminal access channel from the NOAA chart, surfaced by the explorer's tool-returned image pass.}
    \end{minipage}\hfill
    \begin{minipage}[t]{0.46\linewidth}
        \centering
        \includegraphics[height=4.0cm]{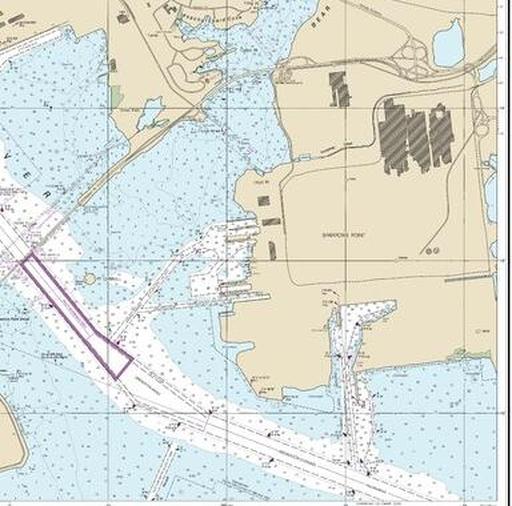}
        \caption*{\footnotesize\emph{Round $t{+}1$ task image} $\mathcal{I}_0$: the curated chart excerpt showing the purple-outlined deep-draft channel reach adjacent to the Seagirt Marine Terminal.}
    \end{minipage}
    \caption{\textbf{Round $t{+}1$ visual artifacts}, produced under the updated $\mathcal{C}_{t+1}$. The graph organizer has higher reasoning and perception step budgets in this round, and the curator anchors the question on a fine-grained channel reach rather than a coarse legend category.}
    \label{fig:case_next}
\end{figure}

\begin{adebox}[colframe=adeteal!85!black,colback=adeteal!2!white]{Round $t{+}1$ Forward (compact)}
\textbf{Seed.} \texttt{entity\_with\_image} pair. Entity \emph{NOAA Nautical Chart 12281: Baltimore Harbor, 57th Edition (November 2018)}, domain \texttt{geography}. The seed proposer now lists the chart's image-visible identifiers explicitly (chart number \texttt{12281}, edition \texttt{57th Ed.}, publication date \texttt{Nov.~2018}, cancellation date \texttt{Mar~6, 2024}) and pairs them with two source URLs that include the original NOAA-hosted PDF.
\textbf{Explorer.} 13 visited URLs over a single phase, surfacing 7 anchors. Anchors include the chart itself, the federally authorized Baltimore Harbor channel network, the USACE maintenance dredging program, the Seagirt Marine Terminal, double-stack intermodal rail, the FY2025 contract opportunity record on \texttt{sam.gov}, and the USACE district news release announcing the FY2025 work.
\textbf{Graph.} Core themes are NOAA charting for Baltimore Harbor, federal navigation channel maintenance, and the Port of Baltimore cargo capacity. Edge structure links Node~1 (Chart 12281) to Node~3 (Baltimore Harbor Federal Navigation Channels) which is \emph{maintained by} Node~4 (USACE program), specialized to Node~7 (FY2025 fiscal year implementation). Reasoning enrichment attaches the derived \emph{average cost per cubic yard} to the edge between the contract value (\$33.5\,M) and the dredge quantity (2.3\,M\,cy).
\textbf{Curator.} The curator emits the dredging task, with difficulty \texttt{expert} and ability profile \texttt{perception\_search\_reasoning}, both raised relative to round $t$.
\end{adebox}

\subsection{Round $t{+}1$ Backward and Update to $\mathcal{C}_{t+2}$}
\label{app:case_next_backward}

The round $t{+}1$ candidate is rolled out by the same policy and scored by the same rubric. The new failure mode is informative.

\begin{adebox}[colframe=adered!85!black,colback=adered!2!white]{Round $t{+}1$ Rubric Scores (overall $2.60$, cognitive friction \texttt{optimal})}
\textbf{Information\_Complexity (5.0).} Multi-step disambiguation among similarly named USACE FY2025 dredging efforts plus official-source verification of contract value and dredge quantity, then cost-per-yard arithmetic.
\textbf{Visual\_Dependency (5.0).} The chart's federal-channel styling and the adjacent terminal label are required to fix the program identity.
\textbf{Verifiability (5.0).} Single objective dollars-and-cents target.
\textbf{Capability\_Requirement (5.0).} Demands chart reading, strategic web retrieval, disambiguation, and arithmetic.
\textbf{Difficulty\_Match ($-3.0$, tag {\small\texttt{[too\_hard]}}).} The task landed \emph{outside} the policy's productive learning zone. The agent never reliably extracted the terminal name and never reached the FY2025 contract record, looping on zoom and visual-search attempts and ending without the required numeric computation.
\textbf{Learning\_Utility (3.0).} The failure is correctable and clearly attributable, so the trace still yields transferable training signal.
\end{adebox}

\begin{adecodebox}[colframe=adegreen!85!black]{Diff $\mathcal{C}_{t+1} \to \mathcal{C}_{t+2}$ (numerical fields)}
explorer.params.max\_nodes\_per\_phase  : 2 $\to$ 1\\
explorer.params.image\_ratio          : 0.40 $\to$ 0.50 \quad{\small(rolled back)}\\
graph\_organizer.complexity.reasoning\_max\_steps   : 6 $\to$ 7\\
graph\_organizer.complexity.perception\_max\_steps  : 5 $\to$ 6
\end{adecodebox}

The {\small\texttt{[too\_hard]}} tag dominates the round's batch-level signal. The optimizer reads this as a request to slow down the explorer (\texttt{max\_nodes\_per\_phase} drops from $2$ to $1$, forcing a deeper traversal of fewer per-phase nodes) and to give the graph organizer more enrichment headroom (both \texttt{reasoning\_max\_steps} and \texttt{perception\_max\_steps} go up by one). The optimizer also \emph{rolls back} the round-$t$ edit on \texttt{explorer.params.image\_ratio} from $0.40$ to $0.50$, since the round-$t{+}1$ failure traces back to insufficient image-bearing evidence per anchor rather than to fetch volume. This rollback illustrates the on-policy character of ODE. Edits are not monotone refinements of a fixed direction. They are responses to whichever failure mode the policy currently exposes, and the optimizer is free to revisit a prior decision once the rollouts under that decision show it was the wrong move.

\subsection{Additional Statistics of ODE-Curated Data}
\label{app:data_stats_extra}

This appendix complements Fig.~\ref{fig:ode_stats} in Section~\ref{sec:data_stats} with the ODE-8B and ODE-30B topic domain donuts and the planned reasoning-step distribution.

\begin{figure}[h]
    \centering
    \begin{subfigure}[t]{0.34\linewidth}
        \centering
        \includegraphics[width=\linewidth]{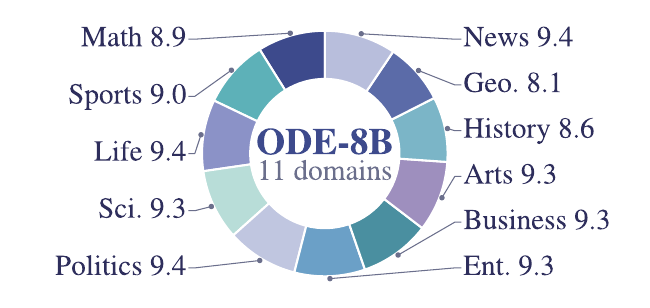}
        \caption{}
        \label{fig:app_domain_8b}
    \end{subfigure}
    \hfill
    \begin{subfigure}[t]{0.34\linewidth}
        \centering
        \includegraphics[width=\linewidth]{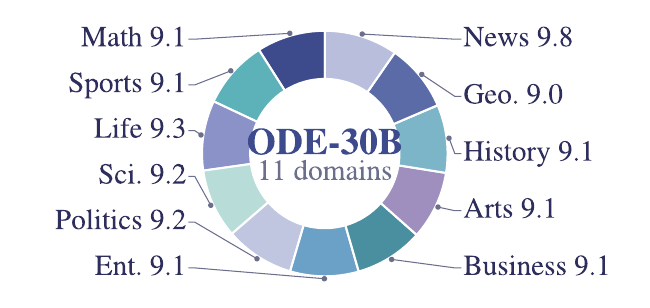}
        \caption{}
        \label{fig:app_domain_rl}
    \end{subfigure}
    \hfill
    \begin{subfigure}[t]{0.30\linewidth}
        \centering
        \raisebox{-0.6em}{\includegraphics[width=\linewidth]{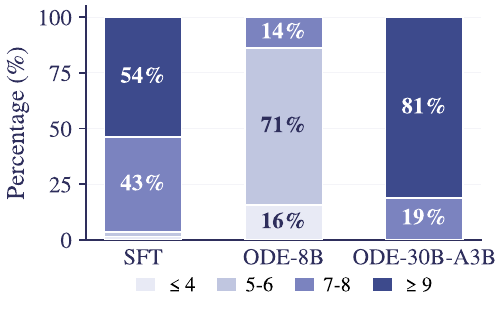}}
        \caption{}
        \label{fig:steps}
    \end{subfigure}
    \caption{\textbf{Additional statistics of ODE-curated data.} Topic domain donuts for the two RL task sets and the planned reasoning-step distribution across the SFT demonstration set and the two RL task sets.}
    \label{fig:data_stats_extra}
\end{figure}

\paragraph{Topical breadth on the RL task sets.} The ODE-8B and ODE-30B shown in Fig.~\ref{fig:data_stats_extra}(a, b) span the same eleven domains as the SFT demonstration set in Fig.~\ref{fig:domain_sft}. ODE-30B confines per-domain shares to a narrow $8.43\%$--$10.03\%$ band with a coefficient of variation of $0.05$, and ODE-8B falls within a comparable band. The forward exploration stage therefore preserves topical coverage across SFT and policy-specific RL data construction, even though the difficulty distribution is allowed to shift between them.

\paragraph{Planned reasoning-step distribution tracks policy capacity.} The reasoning-step buckets in Fig.~\ref{fig:data_stats_extra}(c) read from left to right as a clear depth ladder. ODE-8B concentrates at $5$--$6$ steps with $70.58\%$ of tasks in that bucket, ODE-30B pushes out to $\geq 9$ steps with $81.22\%$, and the SFT demonstration set sits at the deep end with an average of $8.47$ steps inherited from the teacher. The curator's planned-step field therefore tracks each retention's intended trajectory depth, scaling back to shorter plans when the targeted policy cannot sustain long ones and lengthening them when the policy can.

\section{More on Experimental Setup}

\subsection{Data Construction}
\label{app:data}
We use ODE to curate both SFT and RL training data. GPT-5.2~\citep{openai2025gpt52systemcard} is used for most generation, analysis, and
optimization stages, and also serves as the SFT rollout policy; in RL mode, the policy being trained is used for task-verification rollouts. We initialize the evolvable configuration with GPT-5.2 and set the maximum number of evolution steps to 5, with each step using 32 curated tasks and verified traces for rubric-guided configuration updates. We then freeze the selected configuration for large-scale synthesis, yielding 8,855
filtered SFT examples and two RL datasets of 4,000 examples for each model size.

\subsection{Training setup.}
\label{app:train_setup}
We instantiate our agent with two Qwen3-VL backbones: Qwen3-VL-8B-Instruct and Qwen3-VL-30B-A3B-Instruct~\citep{qwen3vl2025}. We first perform SFT on data curated by the SFT mode of ODE. For both
backbones, SFT uses a maximum sequence length of 64k tokens, a global batch size of 64, a learning rate of $2\times10^{-5}$, and 2 training epochs.
Starting from the SFT checkpoints, we further refine the agents with reinforcement learning.
Following~\citet{huang2026visiondeepresearchincentivizingdeepresearchcapability,chen2026opensearchvlopenrecipefrontier}, we use Group Relative
Policy Optimization (GRPO)~\citep{shao2024deepseekmathpushinglimitsmathematical} with the leave-one-out trick~\citep{ahmadian2024basicsrevisitingreinforcestyle}. RL is conducted in our visual-native agent harness with asynchronous SGLang~\citep{zheng2024sglangefficientexecutionstructured} rollouts, sampling 6 responses per prompt. We use a batch size of 96, an actor learning rate of $2\times10^{-6}$, a clip ratio of 0.28, and no KL regularization. All training experiments were conducted on NVIDIA H20 GPUs.

\subsection{Evaluation Setup}
\label{app:eval_setup}
All models and agents are evaluated under the same decoding and interaction budget. We use temperature $0.6$ and top-$p$ $0.95$, and allow at most 50 LLM calls, 8,192 tokens per turn, and
16,000 tokens in total. To ensure consistent assessment across models and settings, we evaluate all
predictions with the same LLM-as-judge verifier.

The verifier computes the terminal answer-level reward for each trajectory. Given a task instance, we
first extract the candidate final answer from the model response. The judge is then given the question,
the reference answer, the extracted candidate answer, and the full model response as auxiliary context.
It is instructed to assess correctness with respect to the reference answer, emphasizing semantic
equivalence rather than exact surface-form matching. The criteria accept paraphrases, standard
abbreviations, harmless formatting variations, entity-name variants, and mathematically equivalent
numeric expressions, while rejecting answers that are ambiguous, incomplete, contradictory, unrelated,
or that mention the reference answer only in an invalid context. The judge returns a structured JSON
object containing a binary correctness decision, an equivalence category, and a short rationale. We use
the binary decision as the final trajectory reward.

\begin{adebox}[colframe=adeblue!85!black,colback=adeblue!2!white]{Prompt for LLM-as-judge} You are a precise final-answer judge.
  Your job is to decide whether the candidate final answer matches the correct answer.
  Be conservative, but grade like a careful human evaluator rather than a brittle string matcher.
  Do not require the candidate to copy the reference wording if it clearly gives the same answer.

  [Question]
  {question}

  [Correct Answer]
  \{reference\_answer\}

  [Candidate Final Answer]
  \{candidate\_answer\}

  [Full Model Response For Context]
  \{full\_response\}

  Task: Determine whether the candidate final answer is correct.

  Decision rules:
  1. Judge the candidate final answer itself, not the quality or completeness of the explanation.
  2. A short answer can still be fully correct. Do NOT require supporting reasoning, derivations, citations, or
extra context.
  3. If the candidate final answer exactly matches the correct answer after normalizing case, whitespace,
simple punctuation, commas, or trailing zeros, you MUST mark it correct.
  4. First identify what kind of answer the question asks for: a named concept/entity, a definition, a purpose/
function, a mechanism/explanation, a date/date range, a number, an identifier, a location, or a list/set of
items.
  5. Accept direct semantic equivalence when the candidate gives the same answer in a different surface form.
  6. Accept term-definition equivalence if the candidate uniquely names the same concept as the reference
definition, or uniquely gives the defining description of the referenced term.
  7. Accept a more specific but non-contradictory answer when it still directly answers the question.
  8. Accept harmless category or wording variation when it preserves the same core meaning.
  9. Accept a correct answer embedded inside a full sentence if the sentence directly answers the question and
is not contradicted by other content.
  10. For event questions, accept a scene description if it clearly identifies the same event.
  11. For cause, reason, or purpose questions, the stated cause must match; a related but different explanation
is wrong.
  12. For broad class questions, a correct subtype is acceptable.
  13. For numbers, dimensions, and formulas, accept mathematically equivalent forms and reasonable decimal
approximations when they clearly refer to the same value.
  14. For dates or date ranges, every required boundary must match. If the reference is only a year, a full
date within that year is acceptable.
  15. For named entities, a different entity is wrong even if it is similar, related, or from the same family.
  16. Accept standard abbreviated company names and obvious person-name variants when they unambiguously
identify the same entity.
  17. For open-set questions signaled by wording such as "some examples" or "for example", accept alternate
valid examples.
  18. For closed-set questions asking for an exact set of items, extra incorrect items or missing required
items are wrong.
  19. If the candidate gives only a related topic, broad discussion, or background context instead of the
answer itself, mark no.
  20. If the candidate answer is ambiguous, hedged, contradictory, missing required specificity, or not an
answer, mark no.
  21. If the candidate says it could not solve the task, refuses to answer, gives no answer, or only shows tool
traces or search queries, mark no.
  22. If the candidate merely mentions the reference phrase inside a negated statement, failure trace, or
quoted query, mark no.

  Calibration examples:
  - Correct: reference='A set of edges without common vertices.' candidate='matching'
  - Correct: reference='Thought experiments.' candidate='used to explore philosophical questions about
perception and reality'
  - Correct: reference='Saxbys coffee shop' candidate='The Saxbys location at the University of
Pennsylvania...'
  - Correct: reference='Barcelona vs Inter Milan Champions League match' candidate='The image shows Lamine
Yamal celebrating during the Barcelona vs Inter Milan Champions League semi-final.'
  - Correct: reference='2025' candidate='April 17, 2025'
  - Correct: reference='A snake' candidate='Gary, a blue pit viper'
  - Correct: reference='Anker Innovations' candidate='Anker'
  - Correct: reference='Clem Delangue' candidate='Clément Delangue'
  - Correct: reference='log(2)/log(3)' candidate='0.6309'
  - Wrong: reference='April 22 - 29' candidate='April 23 to April 29, 2025'
  - Wrong: reference='The HIVE Evo' candidate='The HIVE - Modular Hex Drawers'
  - Wrong: reference='precautionary checks after a gruelling bout' candidate='severe dehydration from weight
cut'
  - Wrong: reference='1500 light-years' candidate='1375 light-years'
  - Wrong: reference='the Ocean's trilogy' candidate='Ocean's Eleven'
  - Wrong: reference='Ex Machina' candidate='About Time'
  - Wrong: reference='Canvas art prints' candidate='giclee prints'
  - Wrong: reference='Martha' candidate='None of the characters... Martha'

  Return ONLY a single-line JSON object with no markdown fences and no extra text:
  \{"correct":"yes"|"no","equivalence":"exact"|"format"|"semantic"|"wrong"|"missing"|"ambiguous","reason":"one
short sentence"\}
\end{adebox}

\subsection{Benchmark Details} \label{app:benchmark_details}
We provide the detailed evaluation benchmark below. Unless otherwise specified, we evaluate on the released
benchmark questions with our unified agent harness and LLM-based answer judge.

\paragraph{MMBC.}
MM-BrowseComp (MMBC)~\citep{li2025mmbrowsecompcomprehensivebenchmarkmultimodal} is a multimodal browsing benchmark designed to test whether agents can retrieve and reason over web evidence that may appear in images or videos rather than text alone. Its questions are hand-crafted to require multi-hop multimodal browsing,
  and each item includes fine-grained reasoning requirements for checking multimodal dependency. We evaluate on
  the released MMBC evaluation set.

\paragraph{HLE-VL.}
HLE-VL is the visual-language subset of Humanity's Last Exam~\citep{phan2025lastexam}, an expert-level academic benchmark with broad subject coverage and questions designed to be unambiguous, verifiable, and difficult to answer through shallow retrieval. We use HLE-VL to measure whether multimodal agents can combine visual interpretation with specialized academic reasoning.

\paragraph{BC-VL.} Introduced by WebWatcher~\citep{geng2026webwatcher}, BC-VL extends BrowseComp-style hard browsing tasks to the visual domain. The benchmark contains long, entity-obfuscated multimodal questions that require cross-modal inference, web search, browsing, and planning rather than direct perception alone. Following prior work, we evaluate on the full BC-VL split.

\paragraph{VDR.}
VDR-Bench~\citep{zeng2026visiondeepresearchbenchmarkrethinkingvisual} evaluates multimodal deep-research agents under long-horizon visual and textual search. It emphasizes multi-turn, multi-entity, and multi-scale evidence gathering, making it especially relevant for testing whether an agent can combine visual retrieval, textual search, and iterative reasoning. We use the test-mini split.

\paragraph{MMSearch.}
MMSearch~\citep{jiang2025mmsearch} evaluates whether large multimodal models can act as multimodal search engines. It contains manually curated queries spanning news and rare-knowledge domains, requiring models to retrieve external evidence rather than answer from parametric knowledge alone. We evaluate on all VQA instances in MMSearch.

\paragraph{MMSearch+.} MMSearch+~\citep{tao2026mmsearchplus} is a provenance-aware multimodal browsing benchmark designed to require fine-grained visual cue extraction, iterative image-text retrieval, and cross-validation under retrieval noise. We evaluate on the single-image subset, following the setting used in prior multimodal deep search evaluations.

\paragraph{SimpleVQA.}
SimpleVQA~\citep{cheng2025simplevqamultimodalfactualityevaluation} evaluates factuality in multimodal question
answering. Its examples focus on short, factual visual questions where the answer should be grounded in
reliable visual or world knowledge. We randomly sample 300 examples for evaluation.

\paragraph{FVQA.}
FVQA~\citep{wang2017fvqafactbasedvisualquestion} is a fact-based visual question answering benchmark where
answering requires external factual knowledge in addition to image understanding. Each question is associated with supporting facts, making it useful for evaluating knowledge-grounded visual reasoning. We randomly
sample 300 examples for evaluation.

Following prior evaluation practice where applicable~\citep{geng2026webwatcher,huang2026visiondeepresearchincentivizingdeepresearchcapability}, we use the
test-mini split of VDR, the full split of BC-VL, all VQA instances in MMSearch, MMBC, HLE-VL, and the single-image subset of MMSearch+. 
For SimpleVQA and FVQA, we randomly sample 300 instances from each benchmark.

\subsection{Train/Evaluation Overlap Audit}
\label{app:overlap_audit}
Seed entities were selected without using benchmark questions, answers, or entity lists. Web exploration also blocked Hugging Face and benchmark source-project URLs. We audited the SFT dataset and both RL datasets against the eight evaluation sets, comprising 2,586 questions and 2,371 images. Table~\ref{tab:overlap_audit} summarizes the checks. All 51 near-duplicate image candidates identified by pHash were false positives upon manual review. Our audit found no evidence of benchmark contamination in the ODE training data.

\begin{table}[htbp]
    \centering
    \small
    \setlength{\tabcolsep}{4pt}
    \caption{Train/evaluation overlap audit across all three training datasets and eight evaluation sets.}
    \label{tab:overlap_audit}
    \begin{tabular}{llrr}
        \toprule
        Check & Method & Candidates & Confirmed \\
        \midrule
        URLs & Exact intersection & 0 & 0 \\
        Questions/answers & Exact + normalized & 0 & 0 \\
        Near-duplicate questions & Jaccard $\geq 0.70$, $\geq 0.80$ & 0 & 0 \\
        Exact images & Decoded-image SHA-256 & 0 & 0 \\
        Near-duplicate images & pHash $\leq 8$ + manual review & 51 & 0 \\
        \bottomrule
    \end{tabular}
\end{table}
\FloatBarrier

\subsection{Agreement with Independent Trace Analyzers}
\label{app:analyzer_agreement}
GPT-5.5 and Claude Opus 4.8 independently rescored all archived ODE-SFT and ODE-RL-8B traces using the same mode-specific rubrics. Table~\ref{tab:analyzer_agreement} reports Pearson correlations with the original GPT-5.2 scores. Overall correlations range from 0.790 to 0.837, supporting agreement in ODE's internal trace diagnostics across analyzers. Agreement across independent analyzers suggests that ODE's trace diagnostics do not depend strongly on the choice of scoring model.

\begin{table}[htbp]
    \centering
    \small
    \caption{Pearson correlations with GPT-5.2 for independent rescoring of archived traces. Dashes mark dimensions not applicable to the corresponding mode.}
    \label{tab:analyzer_agreement}
    \begin{tabular}{lcccc}
        \toprule
        & \multicolumn{2}{c}{GPT-5.5} & \multicolumn{2}{c}{Claude Opus 4.8} \\
        \cmidrule(lr){2-3}\cmidrule(lr){4-5}
        Rubric dimension & SFT & RL-8B & SFT & RL-8B \\
        \midrule
        Overall & 0.816 & 0.837 & 0.790 & 0.814 \\
        Information Complexity & 0.868 & 0.836 & 0.893 & 0.734 \\
        Visual Dependency & 0.832 & 0.827 & 0.786 & 0.800 \\
        Shortcut Leakage & 0.819 & 0.711 & 0.748 & 0.717 \\
        Verifiability & 0.857 & 0.706 & 0.786 & 0.796 \\
        Step Appropriateness & 0.890 & --- & 0.748 & --- \\
        Tool Usage Quality & 0.895 & --- & 0.768 & --- \\
        Capability Requirement & --- & 0.831 & --- & 0.699 \\
        Difficulty Match & --- & 0.725 & --- & 0.696 \\
        Learning Utility & --- & 0.767 & --- & 0.707 \\
        \bottomrule
    \end{tabular}
\end{table}
\FloatBarrier

\subsection{Repeated-Inference Evaluation}
\label{app:inference_variance}
We performed two additional independent inference runs for each final RL checkpoint, giving three runs including the original evaluation reported in Table~\ref{tab:main}. Table~\ref{tab:inference_variance} reports the mean and standard deviation across these three runs. The eight-benchmark average accuracy is $38.83\pm0.17\%$ for ODE-8B-RL and $41.78\pm0.23\%$ for ODE-30B-A3B-RL. For both ODE models, average benchmark performance remains stable across repeated inference runs with the checkpoints held fixed.

\begin{table}[htbp]
    \centering
    \small
    \caption{Accuracy (\%) as mean $\pm$ standard deviation over three inference runs of fixed checkpoints. Avg. summarizes the average accuracy over all eight benchmarks.}
    \label{tab:inference_variance}
    \begin{tabular}{lcc}
        \toprule
        Benchmark & ODE-8B-RL & ODE-30B-A3B-RL \\
        \midrule
        MMBC & $12.05\pm0.45$ & $11.46\pm0.51$ \\
        HLE-VL & $11.21\pm0.61$ & $11.41\pm0.77$ \\
        BC-VL & $41.52\pm0.39$ & $46.69\pm0.81$ \\
        VDR & $20.53\pm0.42$ & $25.60\pm0.69$ \\
        MMSearch & $65.11\pm1.02$ & $69.22\pm0.39$ \\
        MMSearch+ & $24.43\pm0.46$ & $27.45\pm0.69$ \\
        SVQA & $70.55\pm0.39$ & $72.00\pm0.88$ \\
        FVQA & $65.22\pm0.69$ & $70.44\pm0.96$ \\
        \midrule
        Avg. & $38.83\pm0.17$ & $41.78\pm0.23$ \\
        \bottomrule
    \end{tabular}
\end{table}
\FloatBarrier

\section{More on related work}
\label{app:related}

\subsection{Multimodal Web and Deep Search Agents}

Interactive multimodal agents have been studied in two adjacent settings. In visual web navigation,
agents use rendered webpages, screenshots, accessibility trees, or browser actions to complete
real-world web tasks. SeeAct~\citep{zheng2024gpt4visiongeneralistwebagent} studies GPT-4V as a
generalist web agent and highlights the importance of grounding visual plans into executable browser
actions. WebVoyager~\citep{he2024webvoyagerbuildingendtoendweb} builds an end-to-end
multimodal web agent for real websites, and OpenWebVoyager~\citep{he2024openwebvoyagerbuildingmultimodalweb} further explores iterative real-world exploration,
feedback, and optimization for open multimodal web agents. VisualWebArena~\citep{koh2024visualwebarenaevaluatingmultimodalagents} provides a realistic evaluation environment for
web tasks that require both visual perception and browser interaction.

Multimodal deep search agents focus more directly on collecting and verifying external evidence
through search, browsing, and visual tools. MMSearch~\citep{jiang2025mmsearch} and Vision Search Assistant~\citep{zhang2024visionsearchassistant} connect MLLMs to multimodal search pipelines. MMSearch-R1~\citep{wu2025mmsearchr1incentivizinglmmssearch} studies adaptive image and text search through
reinforcement learning. DeepMMSearch-R1~\citep{narayan2025deepmmsearchr1empoweringmultimodalllms} extends this direction with multi-turn query
refinement and crop-conditioned image search. WebWatcher~\citep{geng2026webwatcher} combines
synthetic cold-start trajectories with reinforcement learning for vision-language deep research, and
Vision-DeepResearch~\citep{huang2026visiondeepresearchincentivizingdeepresearchcapability}
scales long-horizon visual-textual search with multiple tools. Together, these works show that multimodal search benefits from richer interaction environments, stronger grounding, and training data that demonstrates multi-step evidence gathering. Our visual-native harness builds on this direction by making the multimodal search workspace explicit: search, browsing, visual manipulation, and computation operate over a shared image bank rather than over isolated tool outputs.

\subsection{Visual Tool Use and Thinking with Images}

A complementary line of work studies how multimodal models can actively manipulate, revisit, or
construct visual information during reasoning. LLaVA-Plus~\citep{liu2023llavapluslearningusetools} integrates external visual tools into multimodal assistants,
while XSkill~\citep{pmlr-v306-jiang26aq} uses past trajectories to guide tool selection on new tasks. V*~\citep{wu2023vguidedvisualsearch} introduces LLM-guided visual search for high-resolution and
visually crowded images, emphasizing selective visual attention and region-level grounding. More
recently, Visual-ARFT~\citep{liu2025visualarft} trains LVLMs to use web
browsing and code-based image manipulation tools through reinforcement fine-tuning, while
DeepEyes~\citep{zheng2025deepeyes} incentivizes interleaved visual and
textual reasoning through reinforcement learning. Thinking with Generated Images~\citep{chern2025thinkinggeneratedimages} explores a related but distinct direction, where models generate
and refine intermediate visual states as part of reasoning.
These works collectively move multimodal reasoning from passive image understanding toward active
visual computation. This trend is closely aligned with multimodal deep search, where useful visual
evidence is often discovered, transformed, or localized over multiple tool calls rather than provided
only in the original input image. This motivates our treatment of intermediate visual observations as
reusable working memory, so that visual evidence discovered or transformed in one step can continue
to support later search and reasoning.

\subsection{Agentic Data Synthesis and Data Evolution}

Synthetic data has become a major driver for training tool-using and agentic language models.
AgentInstruct~\citep{mitra2024agentinstructgenerativeteachingagentic} uses agentic flows to generate
diverse post-training data from raw documents and code. ToolACE~\citep{liu2025toolacewinningpointsllm} constructs function-calling data through self-evolution, multi-agent
dialogue generation, and verification. APIGen-MT~\citep{prabhakar2025apigenmtagenticpipelinemultiturn} generates multi-turn agent data from task
blueprints and simulated human-agent interaction. These methods show that agentic synthesis can
produce scalable supervision for tool use, function calling, and multi-turn interaction.
Recent work has also begun to close the loop between agent behaviour and data construction~\citep{DBLP:journals/corr/abs-2607-07820}.
LoopTool~\citep{zhang2025looptoolclosingdatatrainingloop} uses model-aware probing, label
verification, and error-driven data expansion to improve LLM tool calling. CoEvolve~\citep{yang2026coevolvetrainingllmagents} studies mutual evolution between agent policies and data in
language-agent environments. Similarly, Agent-World~\citep{dong2026agentworldscalingrealworldenvironment} proposes a self-evolving training arena that dynamically synthesizes real-world environments and tasks to identify capability gaps, enabling the continuous co-evolution of agent policies and tool environments. Tongyi DeepResearch~\citep{tongyideepresearchteam2025tongyideepresearchtechnicalreport} demonstrates the importance of
large-scale automatic data synthesis for long-horizon research agents. These works motivate the
broader view that agent training data should adapt to model behaviour, especially when the target
capability involves long-horizon interaction rather than single-turn prediction. ODE extends this
adaptive-data perspective to multimodal deep search, where rollout feedback must capture not only answer correctness but also visual grounding, tool-use structure, and policy-specific learning value.

\section{Limitations}
\label{app:limitations}
This work has three main limitations. First, ODE introduces an additional pre-computation stage before
large-scale training data synthesis. In our experiments, each evolution epoch uses only 32 rollout traces
for feedback, but the repeated generate--rollout--analyze--update loop still consumes wall-clock time
and API resources because it invokes web tools, visual tools, LLM-based analysis, and configuration
optimization. This cost is amortized once the evolved configuration is frozen for large-scale synthesis,
but improving the efficiency of the evolution loop remains important. Future work could study cheaper
proxy feedback, smaller analyzer models, or more autonomous optimization of the data-generation
recipe.
Second, RL training uses a terminal answer-level binary reward from the LLM-as-judge verifier.
Although ODE uses trace-level analysis during data construction to diagnose visual grounding,
tool-use structure, and difficulty match, the policy optimization signal itself is still based on final
answer correctness. Future work could explore evidence-level or step-level rewards that more directly
credit intermediate evidence gathering, image reuse, and tool-use decisions during reinforcement
learning.
Third, this paper instantiates data evolution in multimodal deep search. The core idea of adapting a
data generator from policy rollouts is more general, and may also benefit other agentic settings such as
text-only deep search or GUI agents. Extending ODE
to these domains would require domain-specific harnesses, rubrics, and retention rules, and we leave
this broader generalization to future work.

\section{Broader Impacts}
\label{app:board}
This work may improve multimodal deep search agents for tasks such as shopping assistance, travel
planning, and troubleshooting from user photos, where grounded evidence from images and the web is
useful. It also offers a research insight that multimodal deep search is best understood as a co-design
problem spanning the interaction workspace and the data-construction loop, rather than as a pure
model-scaling problem. In particular, our results suggest that reusable visual evidence and
policy-aware data evolution are both important for training stronger agents, which may inform future
work on multimodal tool use, agent training, and benchmark design.
At the same time, stronger multimodal deep search agents could be misused for large-scale scraping,
privacy-invasive image identification, or generating plausible but unverified summaries. To mitigate
these risks, we emphasize rate-limited tool access, evidence-citation requirements, filtering of
personally identifying images when applicable, and human oversight for high-stakes use cases.

\stopcontents[appendix]


\newpage
\section*{NeurIPS Paper Checklist}

\begin{enumerate}

\item {\bf Claims}
    \item[] Question: Do the main claims made in the abstract and introduction accurately reflect the paper's contributions and scope?
    \item[] Answer: \answerYes{}
    \item[] Justification: The abstract and Section~\ref{sec:method} introduce both contributions claimed in the paper, the visual-native agent harness with image bank reuse and On-policy Data Evolution (ODE), and the experimental gains reported in the abstract are exactly the numbers verified in Section~\ref{sec:method} and the Experiments section across the same eight benchmarks.

\item {\bf Limitations}
    \item[] Question: Does the paper discuss the limitations of the work performed by the authors?
    \item[] Answer: \answerYes{}
    \item[] Justification: We discuss limitations in Appendix~\ref{app:limitations}, including the
  pre-computation cost of ODE, the use of terminal answer-level rewards for RL, and the current
  instantiation of data evolution in multimodal deep search.

\item {\bf Theory assumptions and proofs}
    \item[] Question: For each theoretical result, does the paper provide the full set of assumptions and a complete (and correct) proof?
    \item[] Answer: \answerNA{}
    \item[] Justification: The paper does not contain formal theoretical results, theorems, or proofs; the contributions are an agent harness design, a closed-loop data construction framework, and empirical evaluation.

    \item {\bf Experimental result reproducibility}
    \item[] Question: Does the paper fully disclose all the information needed to reproduce the main experimental results of the paper to the extent that it affects the main claims and/or conclusions of the paper (regardless of whether the code and data are provided or not)?
    \item[] Answer: \answerYes{}
    \item[] Justification: The harness design (nine tools, image bank protocol), the ODE forward and backward stages, the rubric weights, the data construction protocol, the SFT/RL training hyperparameters, and the evaluation protocol with the LLM-as-judge prompt are all described in Section~\ref{sec:method} and Appendices~\ref{app:ode_details}, \ref{app:data}, \ref{app:train_setup}, and \ref{app:eval_setup}, including a fully worked end-to-end ODE round example.

\item {\bf Open access to data and code}
    \item[] Question: Does the paper provide open access to the data and code, with sufficient instructions to faithfully reproduce the main experimental results, as described in supplemental material?
    \item[] Answer: \answerYes{}
    \item[] Justification: We will release the ODE codebase, the visual-native agent harness, the curated SFT and RL data, and trained model checkpoints under an open-source license upon publication. The Appendix already provides the full forward/backward pipeline configuration, prompts, and hyperparameters needed to reproduce the data construction and training.

\item {\bf Experimental setting/details}
    \item[] Question: Does the paper specify all the training and test details (e.g., data splits, hyperparameters, how they were chosen, type of optimizer) necessary to understand the results?
    \item[] Answer: \answerYes{}
    \item[] Justification: Appendices~\ref{app:data}, \ref{app:train_setup}, and \ref{app:eval_setup} specify the SFT/RL data sizes, sequence length, batch size, learning rate, number of epochs, RL algorithm (GRPO with leave-one-out), rollout sampling, decoding parameters, evaluation budget, and the LLM-as-judge protocol used for all reported numbers.

\item {\bf Experiment statistical significance}
    \item[] Question: Does the paper report error bars suitably and correctly defined or other appropriate information about the statistical significance of the experiments?
    \item[] Answer: \answerYes{}
    \item[] Justification: Appendix~\ref{app:inference_variance} reports mean and standard deviation over three independent inference runs of the fixed ODE-8B-RL and ODE-30B-A3B-RL checkpoints across all eight benchmarks. These statistics capture inference-time variability, not training-seed variability. We do not claim statistical significance for the single-run ablations.

\item {\bf Experiments compute resources}
    \item[] Question: For each experiment, does the paper provide sufficient information on the computer resources (type of compute workers, memory, time of execution) needed to reproduce the experiments?
    \item[] Answer: \answerYes{}
    \item[] Justification: Appendix~\ref{app:train_setup} reports the GPU type, optimizer, batch size, learning rate, sequence length, and number of epochs for SFT, the GRPO setup, rollout sampling and clip ratio for RL, and the asynchronous SGLang serving configuration; we additionally describe the evaluation interaction budget (LLM call cap, per-turn and total token caps) in Appendix~\ref{app:eval_setup}. 

\item {\bf Code of ethics}
    \item[] Question: Does the research conducted in the paper conform, in every respect, with the NeurIPS Code of Ethics \url{https://neurips.cc/public/EthicsGuidelines}?
    \item[] Answer: \answerYes{}
    \item[] Justification: The research conforms with the NeurIPS Code of Ethics. The work does not involve human subjects, uses only publicly available benchmarks and pre-trained models under their respective licenses, and we preserve anonymity in all materials submitted for review.

\item {\bf Broader impacts}
    \item[] Question: Does the paper discuss both potential positive societal impacts and negative societal impacts of the work performed?
    \item[] Answer: \answerYes{}
    \item[] Justification: We discuss positive and negative societal impacts in the Appendix~\ref{app:board}
  section, including practical assistance applications, research insights on workspace and data-loop
  co-design, and risks such as scraping, privacy-invasive image identification, and unverified summaries,
  together with mitigations.

\item {\bf Safeguards}
    \item[] Question: Does the paper describe safeguards that have been put in place for responsible release of data or models that have a high risk for misuse (e.g., pre-trained language models, image generators, or scraped datasets)?
    \item[] Answer: \answerYes{}
    \item[] Justification: The released ODE-curated SFT and RL data are derived from publicly accessible web sources and are checked by an LLM judge for verifiability and shortcut leakage; we plan to release the data and model checkpoints under an open-source license, exclude personally identifying images, and accompany the release with a model card that describes the intended scope (multimodal deep search research) and known failure modes.

\item {\bf Licenses for existing assets}
    \item[] Question: Are the creators or original owners of assets (e.g., code, data, models), used in the paper, properly credited and are the license and terms of use explicitly mentioned and properly respected?
    \item[] Answer: \answerYes{}
    \item[] Justification: All used assets, including the Qwen3-VL-8B-Instruct and Qwen3-VL-30B-A3B-Instruct backbones~\citep{qwen3vl2025}, the eight evaluation benchmarks (MMBC, HLE-VL, BC-VL, VDR, MMSearch, MMSearch+, SimpleVQA, FVQA), GRPO~\citep{shao2024deepseekmathpushinglimitsmathematical}, and SGLang~\citep{zheng2024sglangefficientexecutionstructured}, are cited in the main text and appendix and used in accordance with their original licenses and intended research use.

\item {\bf New assets}
    \item[] Question: Are new assets introduced in the paper well documented and is the documentation provided alongside the assets?
    \item[] Answer: \answerYes{}
    \item[] Justification: We introduce three new assets: the visual-native agent harness, the ODE-curated SFT demonstration set, and two ODE-curated RL task sets (ODE-8B and ODE-30B-A3B). Section~\ref{sec:method} and Appendices~\ref{app:ode_details}, \ref{app:data}, and \ref{app:data_stats_extra} document the construction pipeline, prompts, rubrics, sizes, domain coverage, and difficulty distribution; documentation will accompany the public release.

\item {\bf Crowdsourcing and research with human subjects}
    \item[] Question: For crowdsourcing experiments and research with human subjects, does the paper include the full text of instructions given to participants and screenshots, if applicable, as well as details about compensation (if any)?
    \item[] Answer: \answerNA{}
    \item[] Justification: The paper does not involve crowdsourcing or research with human subjects; all data construction and evaluation are performed automatically through LLM-driven curation, judging, and policy rollout.

\item {\bf Institutional review board (IRB) approvals or equivalent for research with human subjects}
    \item[] Question: Does the paper describe potential risks incurred by study participants, whether such risks were disclosed to the subjects, and whether Institutional Review Board (IRB) approvals (or an equivalent approval/review based on the requirements of your country or institution) were obtained?
    \item[] Answer: \answerNA{}
    \item[] Justification: The paper does not involve human subjects research and therefore does not require IRB approval or equivalent review.

\item {\bf Declaration of LLM usage}
    \item[] Question: Does the paper describe the usage of LLMs if it is an important, original, or non-standard component of the core methods in this research? Note that if the LLM is used only for writing, editing, or formatting purposes and does \emph{not} impact the core methodology, scientific rigor, or originality of the research, declaration is not required.
    \item[] Answer: \answerYes{}
    \item[] Justification: LLMs are a core component of our methodology: Qwen3-VL serves as the trained policy, and a frozen LLM (GPT-5.2) acts as the seed proposer, web explorer, graph organizer, task curator, SFT teacher, trace analyzer, configuration optimizer, and final-answer judge. Their roles, prompts, and rubric weights are documented in Section~\ref{sec:method} and Appendices~\ref{app:ode_details}, \ref{app:data}, and \ref{app:eval_setup}.

\end{enumerate}

\end{document}